\documentclass[pdflatex,sn-vancouver,Numbered]{sn-jnl}

\usepackage{amsmath,amssymb,amsfonts}
\usepackage{amsthm}
\usepackage{algorithm}
\usepackage{algpseudocode}
\usepackage{anyfontsize}
\usepackage{mathrsfs}
\usepackage{bm}

\usepackage{float}
\newfloat{algorithm}{tbp}{loa}
\floatname{algorithm}{Algorithm}

\usepackage[utf8]{inputenc}
\usepackage[T1]{fontenc}
\usepackage{txfonts}
\usepackage{makecell}
\usepackage{textcomp}
\usepackage{microtype}

\usepackage{graphicx}
\makeatletter
\def\caption@documentclass{standard}
\makeatother
\usepackage[style=base]{caption}
\usepackage{subcaption}
\usepackage[figuresright]{rotating}
\usepackage{pdflscape}

\usepackage[table]{xcolor}
\usepackage{booktabs}

\usepackage{multirow}
\usepackage{tabularx}
\usepackage{threeparttable}

\usepackage{listings}
\usepackage{tcolorbox}
\usepackage{comment}
\usepackage{pifont}
\usepackage{nicefrac}
\usepackage{lineno}
\usepackage{silence}
\usepackage{url}
\usepackage{hyperref}
\definecolor{darkblue}{rgb}{0.10,0.20,0.65}
\definecolor{darkred}{rgb}{0.70,0.00,0.00}
\definecolor{darkgreen}{rgb}{0.20,0.50,0.20}
\hypersetup{
  colorlinks=true,
  breaklinks=true,
  linkcolor=darkred,
  urlcolor=darkblue,
  anchorcolor=darkblue,
  citecolor=darkblue
}

\newtheorem{theorem}{Theorem}[section]
\theoremstyle{remark}
\newtheorem{remark}[theorem]{Remark}

\begin{document}

\title[]{Active Learning with Bayesian Multi-Fidelity Laplace Neural Operators for Oscillatory Parametric PDEs}

\author[1]{\fnm{Bongseok} \sur{Kim}}\email{kim4853@purdue.edu}
\author[1]{\fnm{Haoyang} \sur{Zheng}}\email{zheng528@purdue.edu}
\author[2]{\fnm{Michael} \sur{Penwarden}}\email{mspenwa@sandia.gov}
\author*[1,3]{\fnm{Guang} \sur{Lin}}\email{guanglin@purdue.edu}

\affil[1]{\orgdiv{School of Mechanical Engineering}, \orgname{Purdue University}, \orgaddress{\street{585 Purdue Mall}, \city{West Lafayette}, \postcode{47907}, \state{IN}, \country{USA}}}
\affil[2]{\orgname{Sandia National Laboratories}, \orgaddress{\city{Albuquerque}, \postcode{87123}, \state{NM}, \country{USA}}}
\affil[3]{\orgdiv{Department of Mathematics}, \orgname{Purdue University}, \orgaddress{\street{150 North University Street}, \city{West Lafayette}, \postcode{47907}, \state{IN}, \country{USA}}}

\abstract{
Surrogate models of parametric dynamical systems are essential for many-query and real-time predictions in engineering applications such as design optimization and digital twins.
However, generating high-fidelity (HF) training data over a broad range of parameters and operating conditions remains computationally expensive.
To address this challenge, we propose a Bayesian multi-fidelity Laplace neural operator (MF-LNO) for uncertainty-aware active learning of oscillatory parametric PDEs.
Specifically, the proposed Bayesian MF-LNO iteratively calibrates the discrepancy between low- and high-fidelity data, where predictive uncertainty guides the adaptive acquisition of informative HF trajectories.
Such predictive uncertainty is quantified via replica-exchange stochastic gradient Langevin dynamics (reSGLD), whose broad posterior exploration enables uncertainty to serve as an error indicator for adaptive HF sample acquisition.
Numerical experiments on the Lorenz system, Duffing oscillator, and beam dynamics demonstrate that uncertainty-guided HF sample acquisition consistently outperforms random sampling, while the proposed Bayesian MF-LNO achieves higher prediction accuracy than MF-DeepONet with predictive uncertainty quantification.
These results demonstrate that Bayesian multi-fidelity LNOs, combined with uncertainty-guided active learning, provide a data-efficient framework for operator learning in engineering dynamical systems.
}


\maketitle

\section{Introduction}
\label{sec_intro}

In computational science and engineering, solving ordinary differential equations (ODEs) and partial differential equations (PDEs) is crucial for modeling complex phenomena across a range of disciplines, including fluid dynamics, structural dynamics, and thermal transport.
Traditional numerical solvers, such as finite element \citep{hughes2003finite} and finite difference methods \citep{leveque2007finite}, provide high-fidelity (HF) solutions but are often computationally prohibitive for many-query or real-time applications.
To overcome these limitations, scientific machine learning~\citep{stevens2020ai} has emerged as a promising paradigm in computational science and engineering, leveraging deep learning to accelerate scientific computing and solve complex scientific problems.
Among these approaches, physics-informed neural networks (PINNs) \citep{raissi2019physics,karniadakis2021physics} have gained considerable attention by directly embedding physical laws (e.g., governing PDEs) into the neural network loss function.
However, since PINNs typically require solving a separate optimization problem for each parameter instance, their applicability to many-query or real-time prediction remains limited.

To address this limitation, operator learning has been developed to learn mappings between infinite-dimensional function spaces using deep neural networks~\citep{lu2021learning}.
A primary application of operator learning is surrogate modeling for parametric PDEs, where solution operators are learned from problem inputs, such as initial conditions, boundary conditions, external forcing, or physical parameters, to the corresponding solutions.
Several architectures, including deep operator networks \citep{lu2021learning, lu2021deepxde}, Fourier neural operators \citep{li2020fourier, kovachki2023neural}, Laplace neural operators (LNO)~\citep{cao2024laplace}, and non-local kernel networks \citep{you2022nonlocal}, etc., have been designed to approximate the mapping between functions. 
Among these, LNOs explicitly model both transient and steady-state responses through a pole-residue formulation in the Laplace domain, providing a unified representation of system poles and frequency-domain responses~\citep{cao2024laplace}. 
Built upon the pole-residue formulation~\citep{hu2016pole}, the LNOs explicitly parameterize system poles and residues, yielding a representation that accommodates oscillatory dynamics and mixed transient-steady-state behavior.

Despite these developments, operator learning generally requires a substantial amount of HF data to construct accurate surrogate models~\citep{lu2022multifidelity}. HF data are typically generated from expensive numerical simulations with fine spatial or temporal resolutions or acquired from high-resolution measurements, making large-scale data collection computationally prohibitive for many engineering applications. To address this challenge, multi-fidelity (MF) modeling combines abundant low-fidelity (LF) data with limited HF data, reducing the need for expensive HF simulations while maintaining predictive accuracy~\citep{peherstorfer2018survey, perdikaris2017nonlinear}. MF approaches utilize correlations between LF and HF data through information fusion and discrepancy modeling to improve prediction accuracy with fewer HF samples~\citep{meng2020composite, howard2023multifidelity}. The MF concept has also been applied to PINNs, where network architectures and optimization settings are treated as different fidelity levels~\citep{penwarden2022multifidelity}. 
In the context of operator learning, MF formulations have been extended to neural operators, where neural operators leverage LF and HF solution data while requiring only a limited amount of HF data~\citep{lu2021learning,lu2022multifidelity}.

While MF modeling reduces the overall cost of HF data acquisition, the selection of additional HF samples remains an important consideration. Uncertainty-guided active learning has recently been incorporated into operator learning to adaptively acquire training data~\citep{winovich2026active}. In parallel, active learning has also been investigated in multi-fidelity modeling using Bayesian latent-variable models and Gaussian process (GP)-based uncertainty estimates~\citep{wu2023disentangled,dhulipala2022active,prentzas2026bayesian}. More recently, Bayesian multifidelity neural operators have combined uncertainty quantification with active learning to adaptively acquire low-fidelity simulation data for mechanics applications~\citep{you2026multi}. 
To the best of our knowledge, however, uncertainty-guided active acquisition of high-fidelity data has not been incorporated into multifidelity neural operator learning for parametric PDEs, although such models are important for many-query and real-time PDE simulations.

% In this work, we propose an uncertainty-guided active learning based on multi-fidelity Laplace neural operators (MF-LNOs) for parametric oscillatory PDEs. Specifically, the proposed framework consists of three components:
% (i) a MF-LNO that leverages a low-fidelity prediction together with parallel linear and nonlinear correction operators to model oscillatory dynamics while efficiently capturing the discrepancies between LF and HF solution operators using limited HF data, where LNOs are adopted for their pole-residue representation, which naturally captures oscillatory, transient, and steady state behavior;
% (ii) replica exchange stochastic gradient Langevin dynamics (reSGLD) to approximate the posterior distribution of the MF-LNO parameters conditioned on the observed data, while employing multiple temperature replicas to improve posterior exploration and thereby provide predictive uncertainty estimates;
% and 
% (iii) an uncertainty-guided active learning strategy that iteratively selects informative HF samples based on the estimated predictive uncertainty. Together, these components enable data-efficient multifidelity operator learning for parametric oscillatory PDEs. 
% Numerical experiments are conducted on representative oscillatory dynamical systems, including challenging cases where LF and HF models exhibit distinct oscillatory characteristics beyond simple linear and nonlinear discrepancies.

In this work, we propose an uncertainty-guided active learning framework for multi-fidelity Laplace neural operators (MF-LNOs) to enable data-efficient learning of solution operators for parametric oscillatory PDEs. The proposed framework leverages abundant low-fidelity (LF) data together with limited high-fidelity (HF) observations to construct a probabilistic approximation of the HF solution operator. Specifically, the MF-LNO augments LF predictions through two parallel correction pathways: a linear operator that captures direct LF-to-HF correlations and a nonlinear Laplace neural operator that learns more complex discrepancies. The latter exploits the pole-residue representation of the LNO to effectively capture differences in oscillatory, transient, and steady-state dynamics between the LF and HF solutions.
To quantify and propagate uncertainty arising from limited HF information, we capitalize on replica-exchange stochastic gradient Langevin dynamics (reSGLD) to approximate the posterior distribution of the MF-LNO parameters conditioned jointly on the available LF and HF data. The resulting posterior predictive uncertainty is used to identify insufficiently resolved regions of the parameter space and acquire informative HF samples. As new HF observations are incorporated, the posterior uncertainty is iteratively updated and propagated through the MF-LNO to guide subsequent HF data acquisition, forming a closed adaptive learning loop. Numerical experiments on representative oscillatory dynamical systems demonstrate the effectiveness of the proposed framework, including challenging cases where the LF and HF models exhibit distinct oscillatory characteristics beyond simple linear and nonlinear discrepancies.
 Numerical experiments are conducted on representative oscillatory dynamical systems, including challenging cases in which the LF and HF models exhibit distinct oscillatory characteristics beyond simple linear and nonlinear discrepancies.

The remainder of this paper is organized as follows. Section~\ref{sec_method} presents the proposed methodology, including the MF-LNO, Bayesian uncertainty quantification using reSGLD, and the uncertainty-guided active learning strategy. Section~\ref{sec_exp} evaluates the proposed method on three representative oscillatory dynamical systems: the Lorenz system, the Duffing oscillator, and a cantilever beam problem with an Euler--Bernoulli beam as the low-fidelity model and a two-dimensional perforated beam as the high-fidelity model. Finally, Section~\ref{sec_discuss} concludes the paper and discusses future research directions.

\section{Proposed methodology}
\label{sec_method}
This section presents the proposed Bayesian MF-LNO for uncertainty-guided active learning. We first review the LNO architecture, followed by the proposed MF-LNO. We then extend the model to a Bayesian formulation using reSGLD for predictive uncertainty quantification and finally present an uncertainty-guided active learning strategy for adaptive HF data acquisition.

\subsection{Laplace Neural Operators}
\label{subsec:method_LNO}

Let $\mathcal{G}: \mathcal{X} \to \mathcal{Y}$ be an operator mapping from an input function $f\in\mathcal{X}$ to an output function $u\in\mathcal{Y}$. In many physical problems, $\mathcal{X}$ might represent boundary or initial conditions (or spatially varying parameters), while $\mathcal{Y}$ corresponds to the solution space of parametric ODEs or PDEs. Our goal is to seek a parametric approximation $\mathcal{G}_{\boldsymbol{\theta}}$ to the true operator $\mathcal{G}$. Specifically, we aim to learn
$$\mathcal{G}_{\boldsymbol{\theta}} : \mathcal{X} \to \mathcal{Y},$$
where ${\boldsymbol{\theta}}$ denotes the neural network parameters. To achieve this, we introduce a neural operator architecture based on the Laplace transform, referred to as the LNO \citep{cao2024laplace}.

Given $f(t) \in \mathcal{X}$, we first map $f$ to a higher-dimensional representation:
\begin{equation*}
v(t)  =  \mathcal{P}\bigl(f(t)\bigr), 
\end{equation*}
where $\mathcal{P}$ is typically a shallow fully-connected neural network or a linear transformation. The lifted representation $v(t)$ resides in an intermediate space for further processing in the Laplace layers. 

Next, we define a nonlinear operator that acts on $v(t)$ in two parts: a kernel integral operator, and a linear bias term. Concretely, for $t \in D$, we have
\begin{equation}\label{eq:LNO_nonlinear}
u(t) 
 = 
\sigma \left(
(\kappa * v)(t)
 + 
\mathbf{W} v(t)
\right)
=
\sigma \left(\int_D
\kappa(t - \tau) 
v(\tau) 
d\tau
 + 
\mathbf{W} v(t)
\right),
\end{equation}
where $\sigma$ is a nonlinear activation function, $\mathbf{W}$ is a linear transformation, $\kappa$ is a integration kernel, and $*$ denotes the convolution integral operation. 
\begin{remark}
    For simplicity, we consider a single nonlinear operator in the present formulation. Specifically, the nonlinear operator can incorporate multiple operators for better generalization. We represent the nonlinear operator applied to a function $ u(t) $ as $\mathcal{N}_i[u](t) = \sigma \left( (\kappa_i * u)(t) + \mathbf{W}_i u(t) \right)$, where $i$ denotes the operations for the $i$-th operator. For iterative applications, the output of the $i$-th operator becomes the input to the $(i+1)$-th operator, which leads to a recursive form with $I$ nonlinear operators: $ u(t) = (\mathcal{N}_I \circ \mathcal{N}_{I-1} \circ \cdots \circ \mathcal{N}_1)[v](t) $. 
\end{remark}

To exploit the structural advantages of frequency-domain analysis, we perform the convolution in the Laplace domain. Let
$
K(s) 
 = 
\mathcal{L}\bigl\{\kappa(t)\bigr\}(s),$ and 
$V(s) 
 = 
\mathcal{L}\bigl\{v(t)\bigr\}(s).
$
Then the convolution can be represented as
\begin{equation}\label{eq:LNO_sdomain}
    \mathcal{L}\Bigl\{(\kappa * v)(t)\Bigr\}(s) 
 = 
K(s) V(s).
\end{equation}

We first represent $V(s)$ in the pole-residue form:
\begin{equation}\label{eq:force_laplace}
V(s)
 = 
\sum_{\ell=-\infty}^{\infty}
\frac{\alpha_{\ell}}{ s - i\omega_{\ell} },
\end{equation}
where $\omega_{\ell} = \ell \omega_1$ denotes the complex frequency, $\omega_1$  is the fundamental complex frequency, and $\alpha_{\ell}$ is the complex amplitudes for each frequency. In practice, the amplitudes are obtained from the fast Fourier transform of $v(t)$. Similarly, we parameterize $\kappa$ in the Laplace domain and represent it in another pole-residue form:
\begin{equation}\label{eq:pole_res_lap}
K_\phi(s)
 = 
\sum_{n=1}^{N}
\frac{\beta_{n}}{s - \mu_{n}},
\end{equation}
where $\mu_n$ ({poles}) and $\beta_n$ ({residues}) are trainable parameters. We denote $\phi$ as the set of trainable parameters $\{\beta_1, \cdots, \beta_N, \mu_1, \cdots, \mu_N\}$. We continue to represent $K_\phi(s) V(s)$ into its partial-fraction decomposition. Multiplying these two terms yields
$$K_\phi(s) V(s)=\sum_{n=1}^{N}
\sum_{\ell=-\infty}^{\infty}
\frac{\beta_{n}\alpha_{\ell}}{ (s - \mu_{n})(s - i\omega_{\ell}) }$$
where each term in $K(s) V(s)$ has two distinct poles $s=\mu_{n}$ and $s=i\omega_{\ell}$. 
For poles at $\mu_n$, by the residue theorem we have
\begin{equation}\label{eq:residue_gamma}
    \gamma_n
 = 
\lim_{s \to \mu_n}
\bigl(s - \mu_n\bigr) U(s)
 = 
\beta_n V(\mu_n),\quad V(\mu_n)
 = 
\sum_{\ell=-\infty}^{\infty}
\frac{\alpha_\ell}{ \mu_n - i\omega_\ell }.
\end{equation}

For terms grouped around the poles at $s=i\omega_{\ell}$, the contribution from these poles is
\begin{equation}\label{eq:residue_lambda}
    \lambda_\ell
 = 
\lim_{s \to i\omega_\ell}
\bigl(s - i\omega_\ell\bigr) U(s)
 = 
\alpha_\ell K(i\omega_\ell),\quad 
K(i\omega_\ell)
 = 
\sum_{n=1}^N
\frac{\beta_n}{ i\omega_\ell - \mu_n }.
\end{equation}

Combining them from $ s = \mu_n $ and $ s = i\omega_\ell $, we yield the corresponding partial-fraction decomposition:
\begin{equation}\label{eq:U_pole_res}
K_\phi(s) V(s)=\sum_{n=1}^N \frac{\gamma_n}{s - \mu_n} + \sum_{\ell=-\infty}^{\infty} \frac{\lambda_\ell}{s - i\omega_\ell}.
\end{equation}

Combining \eqref{eq:residue_gamma}-\eqref{eq:residue_lambda} and applying the inverse Laplace transform to \eqref{eq:U_pole_res} yields:
\begin{equation}\label{eq:u_time_domain}
(\kappa * v)(t)
 = 
\sum_{n=1}^N
\gamma_n \exp(\mu_n t)
 + 
\sum_{\ell=-\infty}^{\infty}
\lambda_\ell 
\exp \bigl(i\omega_\ell t\bigr).
\end{equation}

Lastly, we project $u(t)$ in \eqref{eq:LNO_nonlinear} back to the original output dimension (or ODE/PDE solution space $\mathcal{Y}$) via a local transformation $
\mathcal{G}_{\boldsymbol{\theta}} \bigl(f\bigr)(t)
 = 
\mathcal{Q}\bigl(u(t)\bigr),
$ where $\mathcal{Q}$ is often a neural network or a linear transformation. Together, $\{\mathcal{P},\phi,\mathbf{W},\mathcal{Q}\}$ form the parametric mapping $\mathcal{G}_{\boldsymbol{\theta}}$.

\subsection{Multi-fidelity Laplace neural operator}
\label{subsec:multi_fid_model}

\begin{figure}[htp!]
    \centering
    \begin{subfigure}[t]{1\linewidth}
        \centering
        \includegraphics[width=\linewidth]{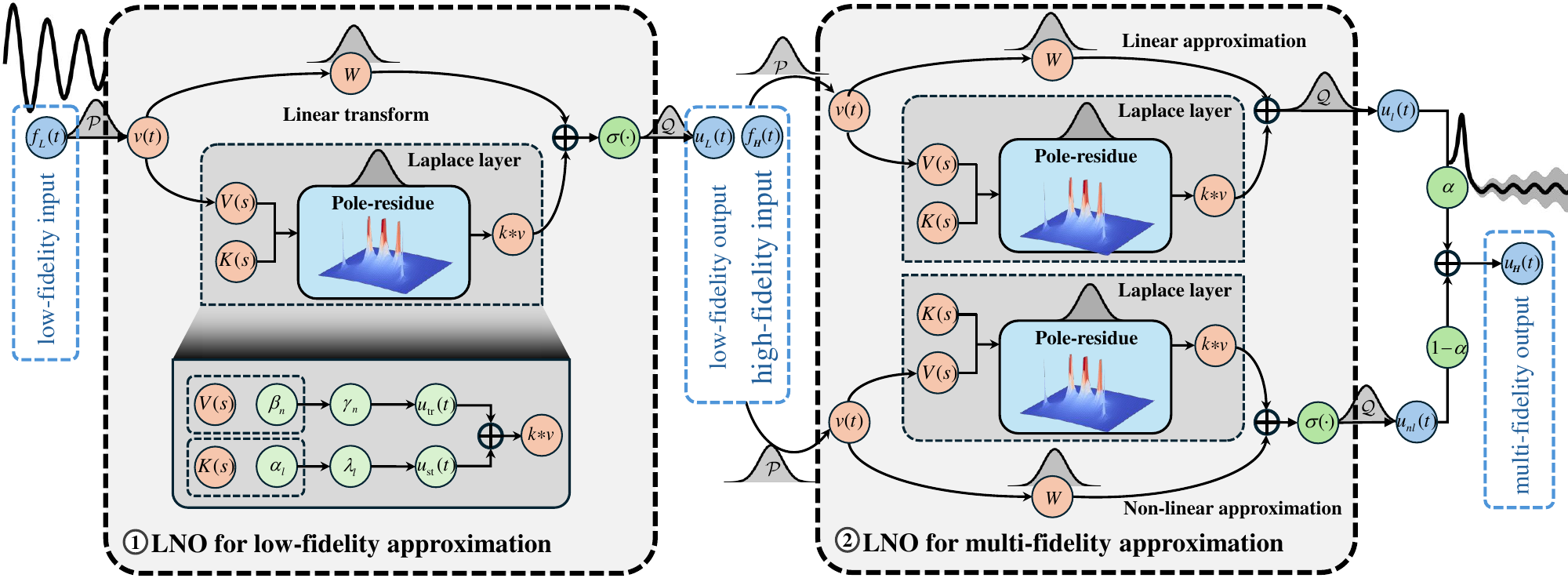}
        % \caption{MF-LNO}
    \end{subfigure}
    \hfill
    \caption{Schematic overview of the proposed MF-LNO, which showcases the integration of LF and HF modeling to achieve efficient MF modeling. The framework begins with \ding{192}an LNO to approximate the mapping from LF inputs to LF outputs, and \ding{193}the MF enhancement involves two parallel LNOs. The linear LNO models the linear relationships between LF and HF data without nonlinear activation functions $\sigma(\cdot)$, while the nonlinear LNO incorporates non-linear corrections through activation functions. Both linear and non-linear LNOs utilize similar architecture, with the addition of trainable weight parameters $\alpha$ to dynamically adjust the contributions from linear and nonlinear operators. Each LNO employs a series of Laplace layers, which leverages kernel convolution and pole-residue parametrization to transform inputs into a feature space optimized for MF approximation. Uncertainty estimates are generated by combining predictions from multiple ensemble models.}
    \label{fig:scheme}
\end{figure}

Building upon the LNO described in Sec.~\ref{subsec:method_LNO}, we propose an MF-LNO that consists of three LNOs: one LNO for learning the LF mapping and two parallel LNOs for modeling the linear and nonlinear relationships between the LF and HF responses.
Figure~\ref{fig:scheme} illustrates the overall architecture.
Given the input functions $f \in \mathcal{X}$, we denote the corresponding LF outputs by $u_L \in \mathcal{Y}_L$ and HF outputs by $u_H \in \mathcal{Y}_H$. For parametric PDEs, the PDE parameters are treated as additional inputs and concatenated with the input function.
In the context of MF surrogate modeling \cite{fernandez2016review}, a commonly utilized relationship between LF and HF data can be expressed as:
\begin{equation}
    u_H(t) = u_L(t) + \mathcal{F}(t) \quad \text{or} \quad u_H(t) = \mathcal{F}(t) \cdot u_L(t),
\end{equation}
where $\mathcal{F}$ represents a correction model that bridges the two fidelity levels. Specifically, $\mathcal{F}(t)$ may take the form of an additive correction term (left) or a multiplicative correction factor (right), which facilitates the transformation of LF outputs $u_L$ into HF outputs $u_H$ while accounting for the discrepancies between them. These corrections can be linear or nonlinear, which depends on the underlying relationships. The primary objective is to construct a surrogate model that learns the mapping from $f$ to $u_H$ using the available datasets:
\begin{equation*}
\{ (f_L^j, u_L^j) \}_{j=1}^{N_L}
\quad
\text{and}
\quad
\{ (f_H^j, u_H^j) \}_{j=1}^{N_H},
\end{equation*}
where $N_L \gg N_H$ in most practical scenarios. LF data are typically abundant but prone to linear or nonlinear biases, leading to reduced accuracy. On the other hand, HF data are more precise but significantly harder to acquire. Directly training an LNO on HF data alone can lead to poor generalization due to the limited size of the HF dataset. Conversely, training the LNO on LF data introduces substantial bias. To mitigate such issues and leverage MF data, we assume that the inter-fidelity relationship between LF and HF data is sufficiently simple to be modeled with limited data. To exploit this, we propose a two-step function mapping process: (1) we first use an LNO, denoted as $\mathcal{G}_{L}$, to learn the mapping $\hat u_L=\mathcal{G}_{L}(f)$. (2) We concatenate the input as $\left[f, \mathcal{G}_{L}(f)\right]$. Then we train a linear LNO (LNO without activation function), $\mathcal{G}_l$, and a non-linear LNO, $\mathcal{G}_{nl}$, to model the relationship from $u_L$ to $u_H$. It should be noted that without the non-linear activation function, the lifting $\mathcal P$, the projection $\mathcal Q$, and $\mathbf{W}$ are clearly linear operations. For the convolution integral in \eqref{eq:U_pole_res}, the computation of $\gamma_n$ and $\gamma_\ell$ depends linearly on $V(s)$, which is derived from the Laplace transform of $v(t)$. This implies that without the non-linear activation function in LNO, it is actually a linear mapping from the input function to the output function. Subsequently, the HF output is modeled as a combination of $\mathcal{G}_l$ and $\mathcal{G}_{nl}$:
\begin{equation}\label{eq:alpha}
\hat u_H  =  \alpha \cdot \mathcal{G}_{l}(f, \mathcal{G}_{L}(f))  +  (1 - \alpha) \cdot \mathcal{G}_{nl}(f, \mathcal{G}_{L}(f)),
\end{equation}

where $\alpha \in [0, 1]$ is a learnable parameter to control the weight of linear and nonlinear correlations. We evaluated various neural network architectures, including fully connected neural networks and convolutional neural networks, to learn the inter-fidelity relationships between LF and HF data. Among these, the LNO structure demonstrated the best performance, which we believe is due to its capacity to handle complex functional relationships.

To learn the function mapping from $f$ to $u_L$, we minimize the relative $L^p$ loss function related to $\mathcal{G}_{L}$:
\begin{equation}\label{eq:loss1}
   \mathcal{L}_L({\boldsymbol{\theta}}_{(1)})  =  \frac{1}{N_L}\sum_{j=1}^{N_L} \frac{\Bigl\|\mathcal{G}_{L}(f_L^j) - u_L^j\Bigr\|_p}{\Bigl\|u_L^j\Bigr\|_p},
   % \Bigl\| \mathcal{G}_{L}(f_L^j) - u_L^j \Bigr\|^2.
\end{equation}
where $\|\cdot\|_p$ denotes the $L{^p}$-norm. Once $\mathcal{G}_{L}$ can approximate the operator well (we can consider a threshold to stop training once the validation loss is small enough, or we can set up the number of epochs to train $\mathcal{G}_{L}$), we continue to minimize the loss function related to $\mathcal{G}_{l}$ and $\mathcal{G}_{nl}$:
\begin{equation}\label{eq:loss2}
   \mathcal{L}_H({\boldsymbol{\theta}}_{(2)})  =  \frac{1}{N_H}\sum_{j=1}^{N_H} \frac{\Bigl\| u_H^j - \hat u_H^j \Bigr\|_p}{\Bigl\| u_H^j \Bigr\|_p} ,
   % \Bigl\| u_H^j - \left[ \alpha \cdot \mathcal{G}_{l}\left(f_H^j, \mathcal{G}_{L}(f_H^j)\right) + (1-\alpha) \cdot \mathcal{G}_{nl}\left(f_H^j, \mathcal{G}_{L}(f_H^j)\right) \right] \Bigr\|^2.
\end{equation}
where $\hat u_H^j = \alpha \cdot \mathcal{G}_{l}\left(f_H^j, \mathcal{G}_{L}(f_H^j)\right) + (1-\alpha) \cdot \mathcal{G}_{nl}\left(f_H^j, \mathcal{G}_{L}(f_H^j)\right) $. The total loss is then expressed as:
\begin{equation}\label{eq:loss3}
\mathcal{L}({\boldsymbol{\theta}})  =  \lambda\mathcal{L}_L({\boldsymbol{\theta}}_{(1)}) + (1-\lambda)\mathcal{L}_H({\boldsymbol{\theta}}_{(2)}),
\end{equation}
where $\{{\boldsymbol{\theta}}_{(1)}, {\boldsymbol{\theta}}_{(2)}, \alpha\}={\boldsymbol{\theta}}$ denotes all learnable parameters in MF-LNO, which includes model parameters of all three LNOs and the learnable weight $\alpha$. The weight coefficient $\lambda$ determines the contribution of the two loss functions.
To improve training efficiency and reduce overfitting, we employ a two-phase training strategy:

\begin{itemize}
    \item \textbf{Phase 1} (LF training in Fig.~\ref{fig:scheme}): We train $\mathcal{G}_L$ to learn the mapping from LF inputs to LF outputs by setting $\lambda=1.0$, thereby optimizing only the LF component.

    \item \textbf{Phase 2} (MF training in Fig.~\ref{fig:scheme}): After $\mathcal{G}_L$ is trained, we freeze ${\boldsymbol{\theta}}_{(1)}$ by setting $\lambda=0.0$ and train the MF component using the concatenated input $\left[f_H,\mathcal{G}_L(f_H)\right]$ to predict the HF outputs.
\end{itemize}

We observe this training process can effectively avoid overfitting during Phase 2. Additionally, it ensures that $\mathcal{G}_L$ is well-trained on LF data before leveraging its outputs to optimize the HF component.

\subsection{Bayesian MF-LNO for uncertainty quantification}

In this section, we propose a Bayesian MF-LNO by incorporating Bayesian inference into the proposed MF-LNO. The resulting predictive uncertainty is later utilized in the uncertainty-guided active learning strategy described in Sec.~\ref{Sec:AL} to iteratively acquire informative HF samples.
Existing Bayesian approaches employ different posterior inference methods to estimate predictive uncertainty. Auto-regressive Gaussian processes \citep{williams1995gaussian, pang2019neural} represent a classical Bayesian approach that estimates posterior distributions to provide both predictive accuracy and uncertainty quantification (UQ). However, their computational complexity grows as $\tilde{O}(N^3)$ with the number of training samples, making them impractical for modeling abundant LF data. Bayesian neural networks alleviate this scalability issue but require efficient posterior inference. For example, \citet{meng2021multi} employed Hamiltonian Monte Carlo (HMC) \citep{neal2011mcmc, tanqi2014stochastic, hoffman2014no} to estimate the posterior distribution of network parameters. However, HMC relies on full-gradient evaluations and can become computationally expensive for large-scale neural networks.

\begin{figure}[htp!]
% {r}{0.40\textwidth}
   \begin{center}
   % \vskip -0.2in
     \includegraphics[width=0.40\linewidth]{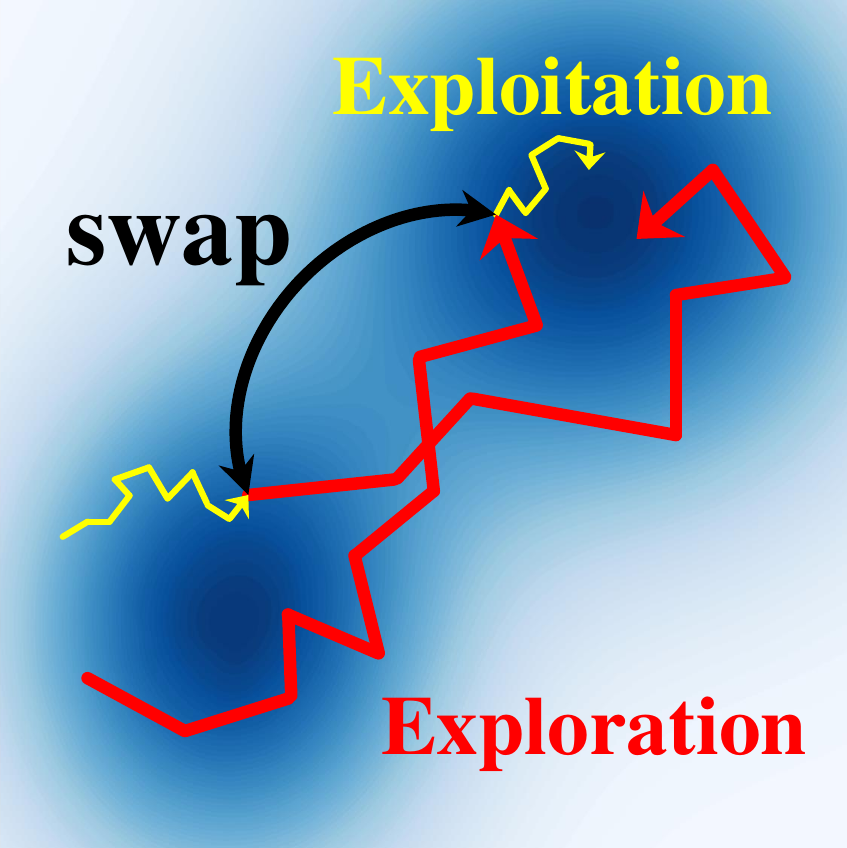}
   \end{center}
   % \vskip -0.0in
    \caption{Sample Trajectory of replica exchange stochastic gradient Langevin dynamics (adapted from \citep{deng2020non, zheng2025exploring}). Yellow lines denote the trajectory of a low-temperature chain (exploitation), and red lines represent a high-temperature chain (exploration). The chain exchange states are according to a swap mechanism. The empirical distribution formed by the yellow trajectory aids in uncertainty quantification.}
    \label{fig:sample_trajectory}
   \vspace{-0.10 in}
\end{figure}

To improve posterior sampling in high-dimensional Bayesian inference, we employ reSGLD. The method combines adaptive moment estimation with replica exchange to balance exploration and exploitation through multiple chains at different temperatures. This enables efficient posterior sampling while improving robustness to multimodal posteriors, noisy gradients, and pathological curvature. The following describes the key components of reSGLD, including SGLD, adaptive drifts, and replica exchange.

First, stochastic gradient Langevin dynamics (SGLD) integrates stochastic gradient descent with Langevin noise to approximate Bayesian posterior sampling in high-dimensional spaces. Given a dataset $\{(f^j, u^j)\}_{j=1}^{N}$, SGLD computes stochastic gradients using mini-batches of data. At iteration $i$, a mini-batch $\mathcal{B}_i \subset \{1, \dots, N\}$ of size $|\mathcal{B}_i|$ is sampled, and the stochastic gradient is $\nabla_{\boldsymbol{\theta}} {\mathcal{L}}({\boldsymbol{\theta}}_i)$, where $\mathcal{L}({\boldsymbol{\theta}}_i)$ is the loss function \eqref{eq:loss3} evaluated at the mini-batch data $\mathcal{B}_i$. The update rule for ${\boldsymbol{\theta}}_i,{\boldsymbol{\theta}}_{i+1}$ with SGLD is then\footnote{We denote here ${\boldsymbol{\theta}}_i$ and ${\boldsymbol{\theta}}_{i+1}$ are neural network parameters of MF-LNO, which includes all parameters in MF-LNO and a learnable weight $\alpha$ mentioned in Section \ref{subsec:multi_fid_model}.}: 
\begin{equation*}
{\boldsymbol{\theta}}_{i+1} = {\boldsymbol{\theta}}_i - \eta_i \nabla_{\boldsymbol{\theta}} \mathcal{L}({\boldsymbol{\theta}}_i) + \sqrt{2 \eta_i \tau} \epsilon_{i}, \quad \epsilon_{i} \sim \mathcal{N}(0, I_d),
\end{equation*}
where $\eta_i$ is the learning rate, $\tau$ is the temperature, and $\epsilon_i$ injects Gaussian noise for exploration. While SGLD is effective in many cases, it struggles with high-dimensional problems characterized by pathological curvature or saddle points. This limitation has motivated the development of adaptive extensions. For example, preconditioning techniques, such as Hessian-based or adaptive preconditioners \citep{girolami2011riemann, li2016preconditioned, deng2019}, scale gradient updates to better navigate ill-conditioned landscapes. Momentum-based methods, including underdamped Langevin Monte Carlo \citep{cheng2018underdamped, zhang2023improved, zheng2024accelerating} and HMC \citep{neal2011mcmc, tanqi2014stochastic, hoffman2014no}, introduce auxiliary variables to improve exploration and escape saddle points. Cyclical MCMC \citep{zhangcyclical}, which periodically varies the step size or noise level, helps balance exploration and exploitation in regions with varying curvature. Noise adaptation strategies, such as adaptive noise scaling \citep{ma2015complete} or annealing, further enhance sampling efficiency. Metropolis-adjusted Langevin algorithm \citep{dwivedi2019log, chewi2021optimal} corrects discretization errors and enables more precise exploration of the parameter space. SGLD with adaptive drifts \citep{kim2022stochastic, ishfaq2024provable} leverages adaptive learning rates based on the first and second moments of the gradients to improve convergence in non-stationary environments. reSGLD \citep{jingdong2, deng2020accelerating} facilitates exploration by swapping states between parallel chains at different temperatures. Referring to the above techniques to escape local modes, we subsequently demonstrate our solution to address such issues when training MF-LNO.

\begin{remark}
    To ensure that SGLD weakly converges to the target distribution, the learning rate $\eta_i$ must satisfy the Robbins-Monro conditions \cite{welling2011bayesian,chen2014stochastic}, specifically $\sum_{i=1}^\infty \eta_i = \infty$ and $\sum_{i=1}^\infty \eta_i^2 < \infty$. This implies that $\eta_i$ decays over time, with $\lim_{i \to \infty} \eta_i = 0$. 
\end{remark}

Next, SGLD with adaptive drifts addresses these challenges by incorporating adaptive moment estimation. The update rule becomes:
\begin{equation*}
{\boldsymbol{\theta}}_{i+1} = {\boldsymbol{\theta}}_i - \eta_i \left( \nabla_{\boldsymbol{\theta}} \mathcal{L}({\boldsymbol{\theta}}_i) + a A_i \right) + \sqrt{2 \eta_i \tau} \epsilon_{i},
\end{equation*}
where $a$ is a scaling factor or weight applied to the adaptive bias term $A_i$, and $A_i = m_i / (\sqrt{V_i} + \rho)$ is an adaptive bias term. The moments $m_i$ and $V_i$ are updated as:  

\begin{equation}\label{eq:moment_estimate}
\begin{aligned}
m_i &= \beta_1 m_{t-1} + (1 - \beta_1) \nabla_{\boldsymbol{\theta}} \mathcal{L}({\boldsymbol{\theta}}_i), \\
V_i &= \beta_2 V_{t-1} + (1 - \beta_2) \left(\nabla_{\boldsymbol{\theta}} \mathcal{L}({\boldsymbol{\theta}}_i)\right)^2,
\end{aligned}
\end{equation}
with $\beta_1, \beta_2$ as exponential decay rates and $\rho$ ensuring numerical stability.

Finally, replica exchange SGLD further enhances exploration by deploying $N_C$ parallel chains at different temperatures $\tau^{(1)} < \cdots < \tau^{(N_C)}$. We denote the chain $n$ follows: 

\begin{equation}\label{eq:resgld}
    {\boldsymbol{\theta}}_{i+1}^{(n)} = {\boldsymbol{\theta}}_i^{(n)} - \eta_i \left( \nabla_{\boldsymbol{\theta}} \mathcal{L}({\boldsymbol{\theta}}_i^{(n)}) + a A_i^{(n)} \right) + \sqrt{2 \eta_i \tau^{(n)}} \epsilon_{i},
\end{equation}
where $A_i^{(n)}$ adapts via chain-specific moments $m_i^{(n)}$ and $V_i^{(n)}$ following \eqref{eq:moment_estimate}. We further consider the strategy to swap chains. Different from a deterministic swap function applied to a two-chain reSGLD, we here employ a deterministic even-odd scheme with gradient-based swaps (see \citep{deng2023non, zheng2024constrained}) to efficiently swap chains in multiple-chain reSGLD. We first introduce the swap mechanism, and then further discuss the update rule for hyper-parameters in the swap mechanism. 

To improve exploration and mixing, the swap mechanism is based on a deterministic even-odd scheme with gradient-based swaps. Unlike conventional Metropolis-Hastings-based replica exchange, the proposed scheme employs gradient-based updates. For adjacent pairs $(n, n+1)$, swaps are attempted alternately within windows of size $\mathbb{W}_*$, which are chosen to avoid insufficient local exploitation and minimize the round-trip time.
The window size
\begin{equation}\label{eq:window}
    \mathbb W_* \geq \left\lceil \frac{\log N_C + \log \log N_C}{-\log (1-\mathbb{S})} \right\rceil
\end{equation}
is a tunable hyper-parameter to minimize round-trip time, and $\mathbb S$ is the target swap rate. A swap between chains $n$ and $n+1$ occurs if both $\mathbb{G}^{(n)} \geq \mathbb{W}_*$ and $\mathbb S^{(n)}_i:=\{\mathcal{L}({\boldsymbol{\theta}}_i^{(n+1)}) + \mathbb{C}_i^{(n)} < \mathcal{L}({\boldsymbol{\theta}}^{(n)})\}$ hold true. Intuitively, the event $\mathbb S^{(n)}_i$ assumes that swaps happen when the higher-temperature chain has explored a state that is better (lower energy) than the state of the lower-temperature chain, even after accounting for the noise in the energy estimates. Here we denote $\mathbb{C}_i^{(n)}$ as a correction buffer, and $\mathbb G^{(n)} = 0$ if chain swap happen. Otherwise, we set $\mathbb G^{(n)} = \mathbb G^{(n)}+1$.

Following the swap mechanism, the correction buffer $\mathbb{C}_i^{(n)}$ is updated to maintain the target swap rate $\mathbb{S}$:
\begin{equation}\label{eq:buffer_correct}
    \mathbb{C}_{i+1}^{(n)} = \mathbb{C}_{i}^{(n)} + \gamma_i \left( \frac{1}{i} \sum_{j=1}^{i} \mathbb{I}_{\mathbb{S}_j^{(n)}} - \mathbb{S} \right),\quad n=1,2,\cdots,N_C-1,
\end{equation}
where $\gamma_i$ is a step size and $\mathbb{I}_{\mathbb{S}_i^{(n)}}$ is an indicator for swap events. 

By iteratively adjusting $\mathbb{C}_{i}^{(n)}$ based on the observed swap rate, the algorithm dynamically balances the exploration of high-temperature chains with the exploitation of low-temperature chains, which leads to efficient sampling from the target posterior distribution.

\subsection{Active learning strategy}
\label{Sec:AL}

The proposed active learning strategy aims to construct an accurate multifidelity model using only a limited number of HF simulations. We assume that a large LF dataset $\mathcal{D}_L$ is available, whereas the initial HF dataset $\mathcal{D}_H^{(0)}$ is small. Starting from $\mathcal{D}_H^{(0)}$, additional HF samples are selected sequentially from a candidate set $\mathcal{C}$ until the prescribed HF budget $N_{\rm HF}$ is reached; see Algorithm~\ref{alg:mf_active_learning}.

At each active learning stage, the Bayesian MF-LNO is trained using the current HF dataset $\mathcal{D}_H$, and the uncertainty score is evaluated over the remaining candidate set $\mathcal{C}_{\rm rem}\subset\mathcal{C}$. Posterior predictive samples are generated by the reSGLD-based Bayesian MF-LNO, where the parameter updates, adaptive moments, and replica exchanges follow Eqs.~\eqref{eq:resgld}, \eqref{eq:moment_estimate}, \eqref{eq:window}, and \eqref{eq:buffer_correct}, respectively.
The acquisition criterion is defined by the temporal average of the posterior standard deviation,
\begin{equation}
\label{eq:AL_score}
\mathcal{A}(\mu)=\frac{1}{N_t}\sum_{j=1}^{N_t}\sigma_{\rm MF}(t_j;\mu),
\end{equation}
where $\sigma_{\rm MF}(t_j;\mu)$ denotes the pointwise posterior standard deviation and $N_t$ is the number of temporal grid points. At each stage, the candidates with the largest values of $\mathcal{A}(\mu)$ are selected, evaluated by the HF solver, and appended to $\mathcal{D}_H$.
\begin{algorithm}
\caption{Uncertainty-guided active learning for Bayesian MF-LNO.}\label{alg:mf_active_learning}
\textbf{Input} LF dataset: $\mathcal{D}_L=\{(\mu_i,u_L^{(i)})\}_{i=1}^{N_L}$;\\
\textbf{Input} initial HF dataset: $\mathcal{D}_H^{(0)}=\{(\mu_i,u_H^{(i)})\}_{i=1}^{N_0}$;\\
\textbf{Input} candidate set: $\mathcal{C}=\{\mu^{(m)}\}_{m=1}^{N_C}$;\\
\textbf{Input} target HF budget: $N_{\rm HF}$; acquisition batch size: $B$. \vspace{0.05 in}

\begin{algorithmic}[1]

\State \textcolor{gray}{$\triangleright$ \textbf{LF-stage training}}
\State Train the LF operator model $\mathcal{G}_L$ using $\mathcal{D}_L$

\State Initialize $\mathcal{D}_H \leftarrow \mathcal{D}_H^{(0)}$
\State Initialize $\mathcal{C}_{\rm rem}\leftarrow
\mathcal{C}\setminus\{\mu:(\mu,u_H)\in\mathcal{D}_H^{(0)}\}$

\While{$|\mathcal{D}_H| < N_{\rm HF}$}

    \State Train or refine Bayesian MF-LNO $\mathcal{G}_{\rm MF}$ using
    $\mathcal{D}_L$, $\mathcal{D}_H$, and $\mathcal{G}_L$ with multi-chain
    reSGLD, including adaptive moment updates and deterministic even-odd
    replica exchanges
    \Comment{By Eqs.~\eqref{eq:moment_estimate}, \eqref{eq:resgld},
    \eqref{eq:window}, and \eqref{eq:buffer_correct}}

    \For{$\mu \in \mathcal{C}_{\rm rem}$}

        \State Draw posterior predictive samples
        $\{u_{\rm MF}^{(s)}(t;\mu)\}_{s=1}^{N_s}$

        \State Compute the pointwise posterior standard deviation
        $\sigma_{\rm MF}(t_j;\mu)$ on $\{t_j\}_{j=1}^{N_t}$

        \State Compute the uncertainty indicator
        $\mathcal{A}(\mu)=\frac{1}{N_t}
        \sum_{j=1}^{N_t}\sigma_{\rm MF}(t_j;\mu)$
        \Comment{By Eq.~\eqref{eq:AL_score}}

    \EndFor

    \State Select $\mathcal{B}_{\rm new}\subset\mathcal{C}_{\rm rem}$
    of size $B$ with the largest values of $\mathcal{A}(\mu)$

    \State Evaluate the HF solver at each
    $\mu\in\mathcal{B}_{\rm new}$

    \State Update
    $\mathcal{D}_H \leftarrow
    \mathcal{D}_H \cup
    \{(\mu,u_H(\mu)):\mu\in\mathcal{B}_{\rm new}\}$

    \State Update
    $\mathcal{C}_{\rm rem}\leftarrow
    \mathcal{C}_{\rm rem}\setminus\mathcal{B}_{\rm new}$

\EndWhile

\State Train the final Bayesian MF-LNO using
$\mathcal{D}_L$ and $\mathcal{D}_H$

\end{algorithmic}

{\textbf{Output}} Final Bayesian multifidelity model
$\mathcal{G}_{\rm MF}$ and HF dataset $\mathcal{D}_H$
\end{algorithm}

\section{Numerical experiments}
\label{sec_exp}

In this section, we evaluate the proposed uncertainty-guided active learning strategy with Bayesian MF-LNO on three benchmark problems: the parametric Lorenz system, the Duffing oscillator, and a composite beam dynamics problem involving a high-fidelity two-dimensional plane-stress elastodynamic model and a low-fidelity one-dimensional Euler--Bernoulli beam model. We compare the proposed method with four baselines: LF-LNO, HF-LNO, MF-LNO with random HF sample acquisition, and MF-DeepONet~\cite{lu2022multifidelity} with random HF sample acquisition. These comparisons evaluate both the effectiveness of uncertainty-guided HF sample acquisition and the performance of the proposed multifidelity operator architecture. All experiments are implemented in \texttt{PyTorch} and conducted on a desktop equipped with an \texttt{AMD Ryzen Threadripper PRO 5955WX} CPU, an \texttt{NVIDIA GeForce RTX 4090} GPU, and 128~GB of DDR4 memory.

\subsection{Experiment 1: The Lorenz system}
\label{Sec:ex1}

The Lorenz system is a classic example of a chaotic dynamical system, originally derived as a simplified model of atmospheric convection. It is governed by the following set of three coupled nonlinear ODEs:
\begin{equation}\label{eq:lorenz}
    \begin{aligned}
        \dot{u} &= \sigma (y - u) + f(t), \\
        \dot{y} &= u (\rho - z) - y, \\
        \dot{z} &= u y - \beta z,
    \end{aligned}
\end{equation}
where $u(t)$, $y(t)$, and $z(t)$ represent the system state variables, $\sigma$ characterizes the ratio of momentum diffusivity to thermal diffusivity, $\rho$ describes the strength of thermal convection, and $\beta$ is a geometric factor associated with the convection cells. The Lorenz system is well known for its strong sensitivity to parameters and initial conditions, which makes it a useful benchmark for nonlinear multifidelity operator learning.

For this problem, we consider a parametric setting in which both the forcing amplitude and the Lorenz parameter vary. We fix $\sigma=10.0$ and $\beta=\frac{8}{3}$, and set the initial conditions to $u(0)=1.0$ and $y(0)=z(0)=0$. The input forcing is given by $f(t)=(0.2+5a)\sin(2\pi t)$ for $t\in[0,20]$, where the amplitude parameter and the Lorenz parameter are sampled from the two-dimensional parameter space $(a,\rho)$ with $a\in[0.01,1.0]$ and $\rho\in[5.0,8.0]$. For each parameter pair $(a,\rho)$, the HF trajectory is obtained by numerically solving Eq.~\eqref{eq:lorenz} on a fine temporal grid and taking $u(t)$ as the target output. The resulting HF inputs and outputs are discretized into 2048 time points, so that $f_H^j,u_H^j\in\mathbb{R}^{2048}$.

The LF dataset is constructed from the same family of trajectories on a coarser temporal grid with 512 time points. To introduce a prescribed discrepancy between the two fidelity levels, the LF output is defined as $u_L(t)=u_H(t)(1+0.3\sin t)+1.5+0.0225\,t$. Accordingly, the LF inputs and outputs are represented by $f_L^j,u_L^j\in\mathbb{R}^{512}$. The training candidate set consists of a $10\times10$ parameter grid, yielding 100 LF/HF parameter instances, from which only a small number of HF samples are selected for multifidelity training.
The neural network architectures used for the MF-LNO and MF-DeepONet are summarized in Table~\ref{tab:mf_operator_network_specs_lorenz}.

\begin{table}[htbp!]
\footnotesize
\renewcommand{\arraystretch}{1.1}
\setlength{\tabcolsep}{6pt}
\centering
\caption{Neural network specifications of the multi-fidelity operator models for the 1D Lorenz problem.}
\label{tab:mf_operator_network_specs_lorenz}

\begin{tabular}{lcc}
\hline
 & MF-LNO & MF-DeepONet \\
\hline

Hidden layers
&
LF: $[128]$, HF: $[2]$
&
\makecell[c]{Branch: $[32,32,32]$\\
Trunk: $[32,32,32]$}
\\

Width
&
$(3,8)$
&
$(32,32)$
\\

Activation
&
Sine
&
GELU
\\

LF parameters
&
$803$
&
$39{,}201$
\\

HF linear parameters
&
$176$
&
$203{,}041$
\\

HF nonlinear parameters
&
$176$
&
$203{,}041$
\\

\hline

Total parameters
&
$\mathbf{1{,}156}$
&
$\mathbf{445{,}284}$

\\
\hline
\end{tabular}
\end{table}

\begin{figure}[htp!]
    \centering
    \includegraphics[width=0.8\linewidth]{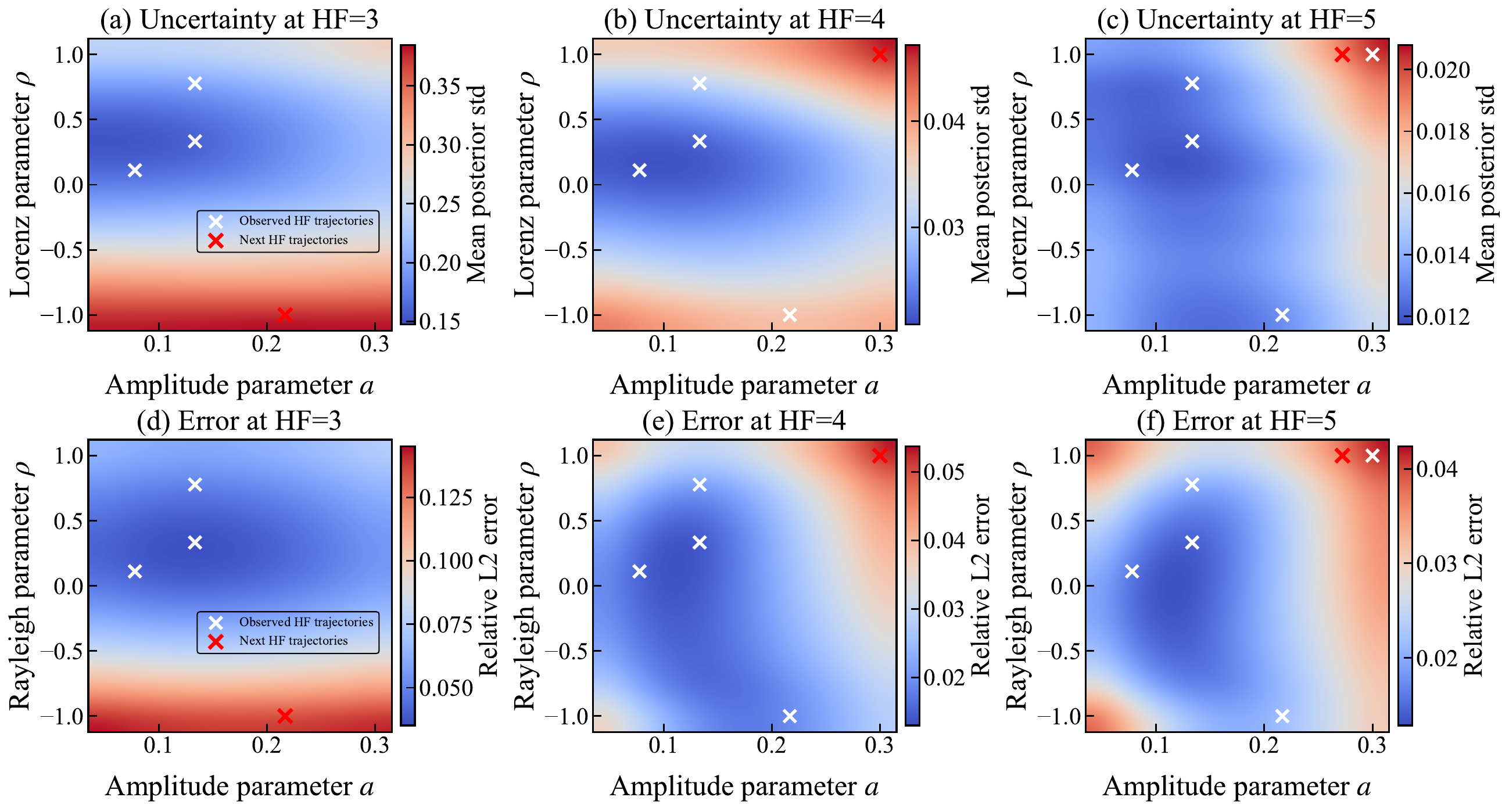}
    \caption{
Uncertainty-guided HF sample acquisition based on reSGLD in the $(a,\rho)$ parameter space. The top row shows the posterior standard deviation, and the bottom row shows the corresponding relative $L_2$ error. Red crosses denote the selected HF samples. The close agreement between the uncertainty and error distributions demonstrates that the posterior uncertainty provides a reliable criterion for HF sample selection.
}
    \label{fig:lorenz_linear}
\end{figure}

To assess whether the posterior uncertainty can reliably identify informative HF samples, Fig.~\ref{fig:lorenz_linear} compares the uncertainty distribution with the corresponding prediction error over the $(a,\rho)$ parameter space. The close agreement between the two distributions indicates that the posterior uncertainty serves as a reliable surrogate for the local HF prediction error, leading to progressive selection of informative HF samples.

\begin{figure}[htp!]
    \centering
    \includegraphics[width=0.8\linewidth]{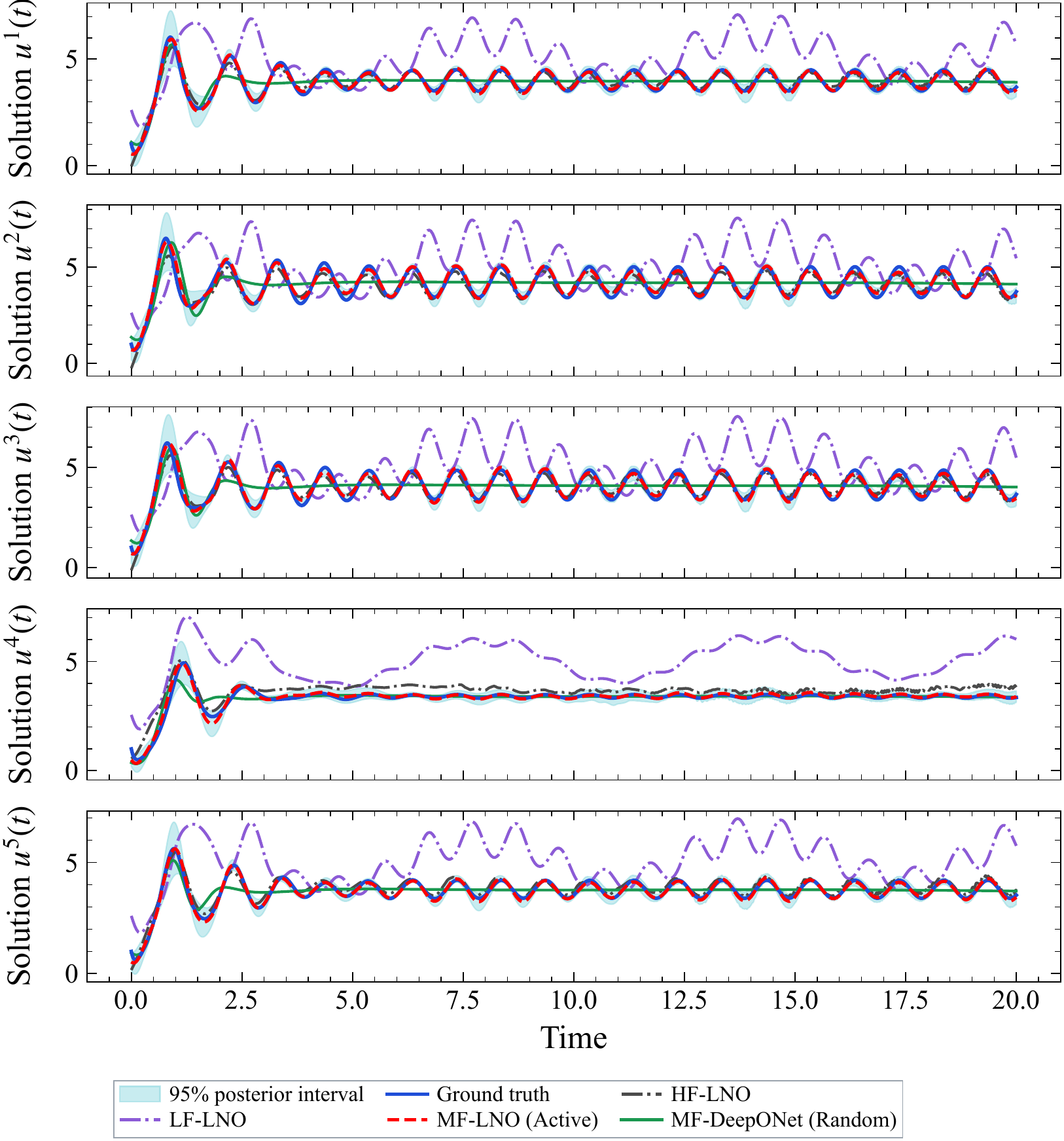}
    \caption{\textbf{1D Lorenz system:} Representative predictions on test samples using nine HF trajectories for multifidelity learning. The blue solid line denotes the ground truth, the red dashed line denotes MF-LNO (Active), the purple dash-dotted line denotes LF-LNO, the gray dash-dotted line denotes HF-LNO, the green solid line denotes MF-DeepONet (Random), and the cyan shaded region denotes the 95\% posterior uncertainty interval of MF-LNO (Active). MF-LNO (Active) accurately reproduces the oscillatory trajectories with only nine HF samples, whereas MF-DeepONet (Random) fails to capture the oscillatory behavior.}
    \label{fig:solution_ex1}
\end{figure}

\begin{figure}[htp!]
    \centering
    \begin{subfigure}[t]{0.4\linewidth}
        % \centering
        \includegraphics[width=\linewidth]{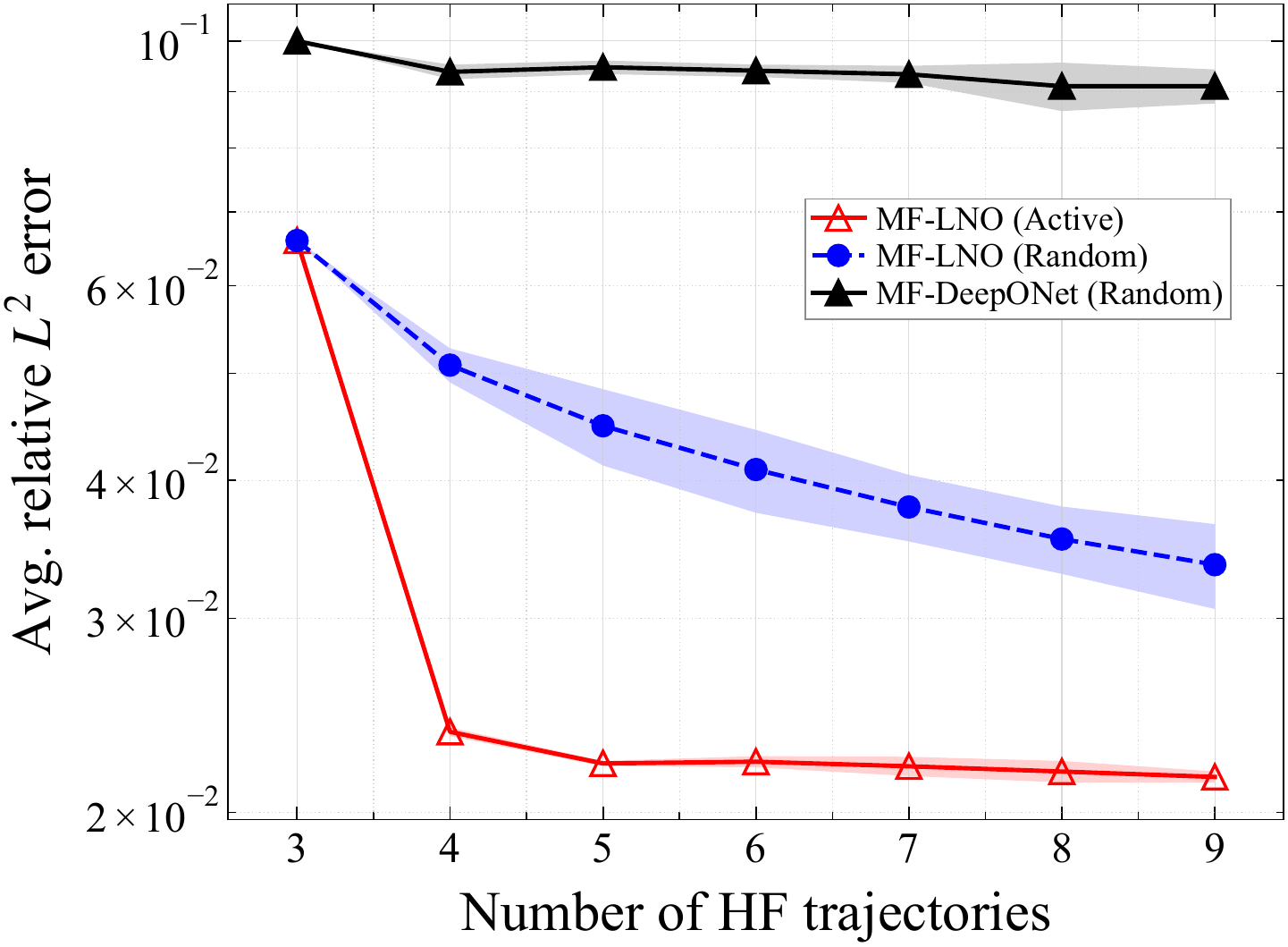}
        \caption{Avg. $L^2$ relative error}
    \end{subfigure}
    % \hfill
    \begin{subfigure}[t]{0.4\linewidth}
        \centering
        \includegraphics[width=\linewidth]{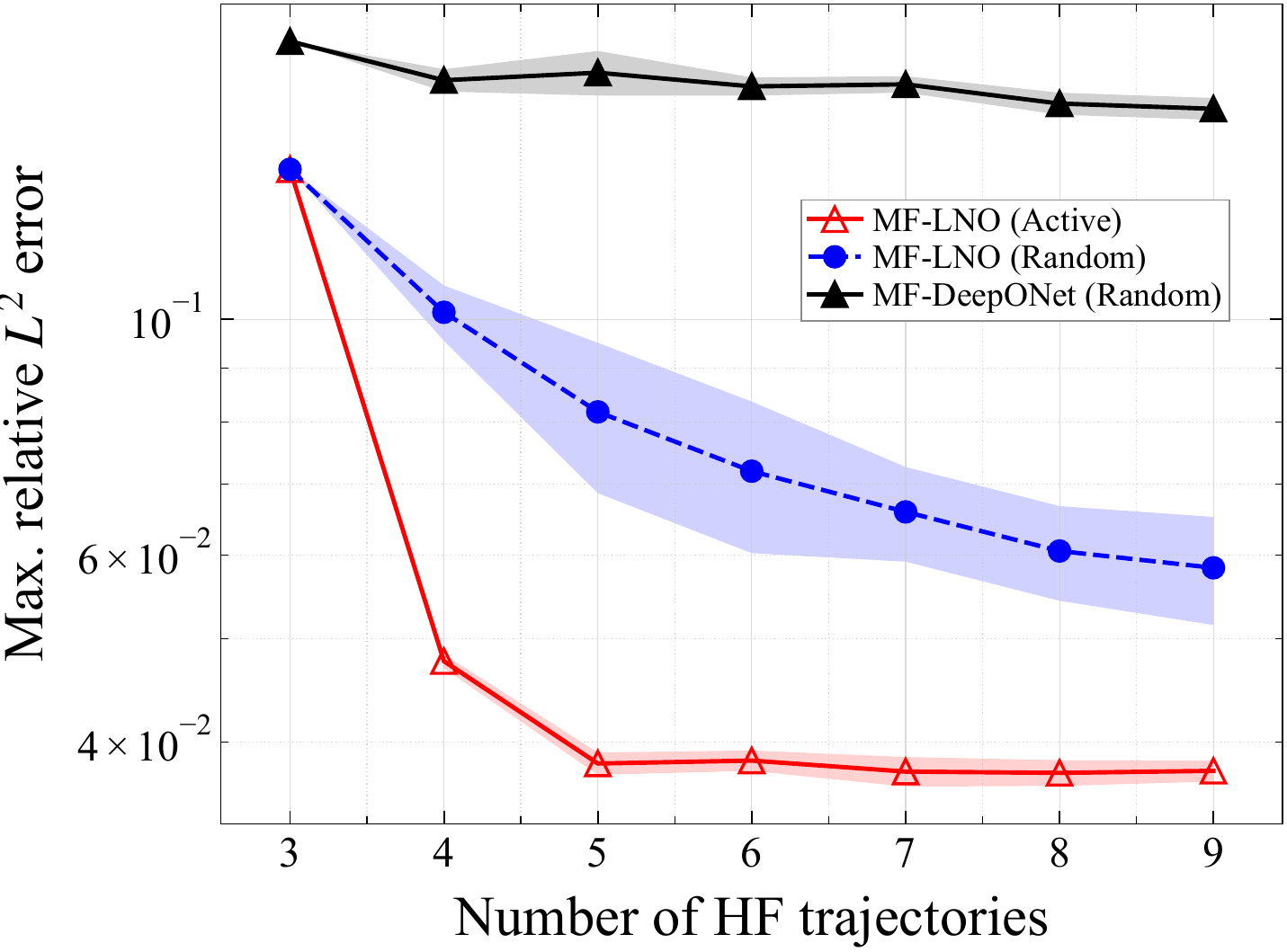}
        \caption{Max. $L^2$ relative error}
    \end{subfigure}
    \caption{\textbf{1D Lorenz system:} Error comparison across five random seeds as the number of HF trajectories increases. The left panel shows the average relative \(L_2\) error, and the right panel shows the maximum relative \(L_2\) error. MF-LNO (Active) employs uncertainty-guided HF trajectory acquisition, whereas MF-LNO (Random) and MF-DeepONet (Random) use random HF trajectory acquisition. All methods are initialized with the same three HF trajectories, and the shaded bands indicate the standard deviation over the five seeds.}
    \label{fig:lorenz_error_comparison}
\end{figure}

\begin{table}[htbp!]
\centering
\footnotesize
\caption{Error comparison for the five representative test trajectories shown in Fig.~\ref{fig:solution_ex1}. MF-LNO (Active) is trained using high-fidelity samples selected by the proposed active learning strategy, whereas MF-DeepONet (Random) is trained using randomly selected high-fidelity samples. All error metrics for MF-LNO (Active) are computed from the posterior mean prediction corresponding to the 95\% posterior interval shown in Fig.~\ref{fig:solution_ex1}. Smaller errors are highlighted in bold for each metric within each trajectory.}
\label{tab:trajectory_error_comparison}
\begin{tabular}{llccc}
\hline
Trajectory & Model & Relative \(L^2\) error & RMSE & MAE \\
\hline
\multirow{4}{*}{\makecell[l]{Fig.~\ref{fig:solution_ex1}: Solution $u^{1}(t)$\\ ($a=0.67,\ \rho=7.00$)}}
& MF-LNO (Active) & \(\mathbf{1.9369\times10^{-2}}\) & \(\mathbf{7.6992\times10^{-2}}\) & \(\mathbf{5.8753\times10^{-2}}\) \\
& MF-DeepONet (Random)     & \(9.6066\times10^{-2}\)          & \(3.8186\times10^{-1}\)          & \(3.3102\times10^{-1}\) \\
& LF-LNO          & \(4.2171\times10^{-1}\)          & \(1.6763\times10^{0}\)          & \(1.3305\times10^{0}\) \\
& HF-LNO          & \(4.1658\times10^{-2}\)          & \(1.6559\times10^{-1}\)          & \(1.2931\times10^{-1}\) \\

\hline
\multirow{4}{*}{\makecell[l]{Fig.~\ref{fig:solution_ex1}: Solution $u^{2}(t)$\\ ($a=0.98,\ \rho=7.72$)}}
& MF-LNO (Active) & \(\mathbf{3.5011\times10^{-2}}\) & \(\mathbf{1.4749\times10^{-1}}\) & \(\mathbf{1.1174\times10^{-1}}\) \\
& MF-DeepONet (Random)     & \(1.4495\times10^{-1}\)          & \(6.1061\times10^{-1}\)          & \(5.4095\times10^{-1}\) \\
& LF-LNO          & \(4.2847\times10^{-1}\)          & \(1.8050\times10^{0}\)          & \(1.4384\times10^{0}\) \\
& HF-LNO          & \(6.6707\times10^{-2}\)          & \(2.8101\times10^{-1}\)          & \(2.2437\times10^{-1}\) \\

\hline
\multirow{4}{*}{\makecell[l]{Fig.~\ref{fig:solution_ex1}: Solution $u^{3}(t)$\\ ($a=0.97,\ \rho=7.38$)}}
& MF-LNO (Active) & \(\mathbf{3.1650\times10^{-2}}\) & \(\mathbf{1.2987\times10^{-1}}\) & \(\mathbf{9.8376\times10^{-2}}\) \\
& MF-DeepONet (Random)     & \(1.3529\times10^{-1}\)          & \(5.5513\times10^{-1}\)          & \(4.9046\times10^{-1}\) \\
& LF-LNO          & \(4.4558\times10^{-1}\)          & \(1.8284\times10^{0}\)          & \(1.4508\times10^{0}\) \\
& HF-LNO          & \(5.8180\times10^{-2}\)          & \(2.3873\times10^{-1}\)          & \(1.9041\times10^{-1}\) \\

\hline
\multirow{4}{*}{\makecell[l]{Fig.~\ref{fig:solution_ex1}: Solution $u^{4}(t)$\\ ($a=0.11,\ \rho=5.35$)}}
& MF-LNO (Active) & \(\mathbf{2.6225\times10^{-2}}\) & \(\mathbf{8.8293\times10^{-2}}\) & \(\mathbf{5.6369\times10^{-2}}\) \\
& MF-DeepONet (Random)     & \(6.6184\times10^{-2}\)          & \(2.2282\times10^{-1}\)          & \(1.1782\times10^{-1}\) \\
& LF-LNO          & \(5.3758\times10^{-1}\)          & \(1.8099\times10^{0}\)          & \(1.6642\times10^{0}\) \\
& HF-LNO          & \(1.1028\times10^{-1}\)          & \(3.7128\times10^{-1}\)          & \(3.3399\times10^{-1}\) \\

\hline
\multirow{4}{*}{\makecell[l]{Fig.~\ref{fig:solution_ex1}: Solution $u^{5}(t)$\\ ($a=0.60,\ \rho=6.41$)}}
& MF-LNO (Active) & \(\mathbf{1.7714\times10^{-2}}\) & \(\mathbf{6.6770\times10^{-2}}\) & \(\mathbf{5.1408\times10^{-2}}\) \\
& MF-DeepONet (Random)     & \(8.7661\times10^{-2}\)          & \(3.3042\times10^{-1}\)          & \(2.7903\times10^{-1}\) \\
& LF-LNO          & \(4.6047\times10^{-1}\)          & \(1.7356\times10^{0}\)          & \(1.4033\times10^{0}\) \\
& HF-LNO          & \(4.6894\times10^{-2}\)          & \(1.7675\times10^{-1}\)          & \(1.4261\times10^{-1}\) \\

\hline
\end{tabular}
\end{table}

\begin{table}[htbp!]
\centering
\footnotesize
\caption{Comparison of the average and maximum relative \(L^2\) errors over all test trajectories for the chaotic parametric Lorenz system with different numbers of high-fidelity (HF) trajectories. MF-LNO (Active) acquires HF trajectories using the proposed active learning strategy based on posterior uncertainty, whereas MF-LNO (Random) and MF-DeepONet (Random) employ random HF trajectory acquisition. All methods are initialized with the same three HF trajectories. The smallest error in each row is highlighted in bold.}
\label{tab:figure4_error_comparison_compact}
\begin{tabular}{clccc}
\hline
\makecell[c]{Number of HF \\ trajectories}
& Metric
& \makecell[c]{MF-LNO \\ (Active)}
& \makecell[c]{MF-LNO \\ (Random)}
& \makecell[c]{MF-DeepONet \\ (Random)} \\
\hline
\multirow{2}{*}{3}
& Avg. error
& \(\mathbf{7.1156\times10^{-2}}\)
& \(\mathbf{7.1156\times10^{-2}}\)
& \(2.2297\times10^{-1}\) \\
& Max error
& \(\mathbf{8.9996\times10^{-2}}\)
& \(\mathbf{8.9996\times10^{-2}}\)
& \(3.4380\times10^{-1}\) \\
\hline
\multirow{2}{*}{5}
& Avg. error
& \(\mathbf{1.6919\times10^{-2}}\)
& \(5.8814\times10^{-2}\)
& \(1.9942\times10^{-1}\) \\
& Max error
& \(\mathbf{4.3242\times10^{-2}}\)
& \(7.2346\times10^{-2}\)
& \(3.5335\times10^{-1}\) \\
\hline
\multirow{2}{*}{7}
& Avg. error
& \(\mathbf{1.4980\times10^{-2}}\)
& \(5.0723\times10^{-2}\)
& \(2.0133\times10^{-1}\) \\
& Max error
& \(\mathbf{3.2625\times10^{-2}}\)
& \(7.0334\times10^{-2}\)
& \(3.5677\times10^{-1}\) \\
\hline
\multirow{2}{*}{9}
& Avg. error
& \(\mathbf{1.4653\times10^{-2}}\)
& \(4.5934\times10^{-2}\)
& \(2.0171\times10^{-1}\) \\
& Max error
& \(\mathbf{3.1769\times10^{-2}}\)
& \(6.6709\times10^{-2}\)
& \(3.5582\times10^{-1}\) \\
\hline
\end{tabular}
\end{table}

Using only nine HF trajectories, the proposed Bayesian MF-LNO accurately reproduces the oscillatory dynamics while providing well-calibrated predictive uncertainty. As shown in Fig.~\ref{fig:solution_ex1}, the proposed method closely matches the ground-truth trajectories, whereas the HF-LNO, LF-LNO, and MF-DeepONet exhibit noticeably larger prediction errors. This qualitative observation is quantitatively confirmed in Table~\ref{tab:trajectory_error_comparison}, where the proposed method consistently achieves the lowest relative $L_2$ error, RMSE, and MAE across all representative test trajectories. As additional HF trajectories are acquired, the prediction errors decrease consistently, with the proposed uncertainty-guided acquisition achieving substantially faster error reduction than random acquisition, as shown in Fig.~\ref{fig:lorenz_error_comparison}. The same trend is reflected in Table~\ref{tab:figure4_error_comparison_compact}, where the proposed Bayesian MF-LNO consistently attains the lowest average and maximum relative $L_2$ errors for all HF sample sizes.

\begin{figure}[htp!]
    \centering
    \begin{subfigure}[t]{0.295\linewidth}
        \centering
        \includegraphics[width=\linewidth]{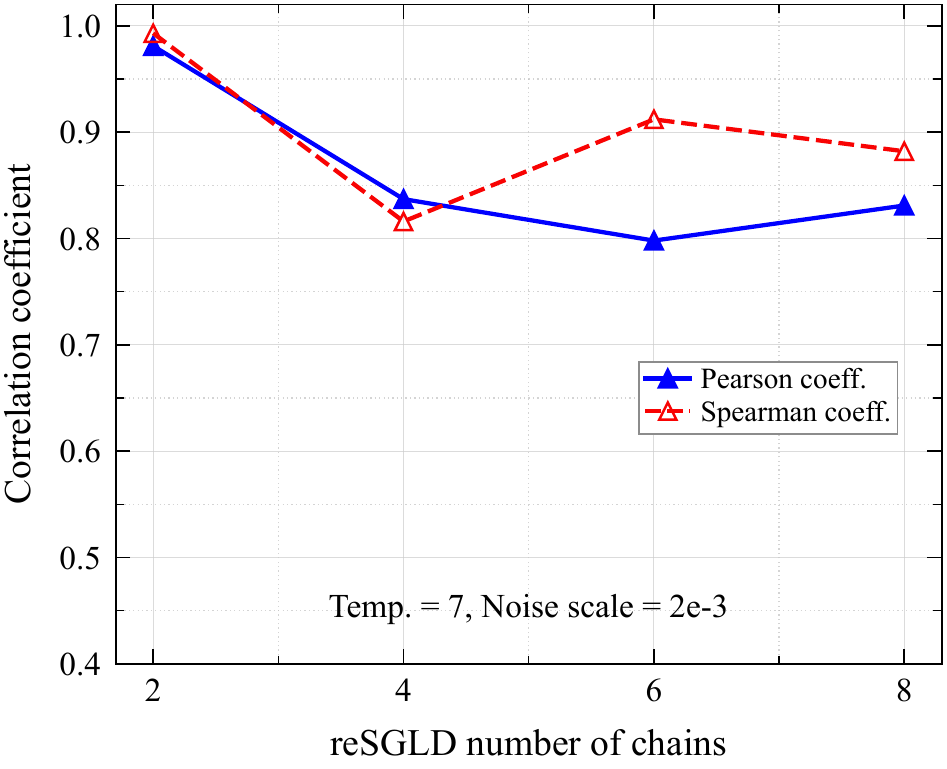}
        \caption{reSGLD chains}
    \end{subfigure}
    % \hfill
    \begin{subfigure}[t]{0.295\linewidth}
        \centering
        \includegraphics[width=\linewidth]{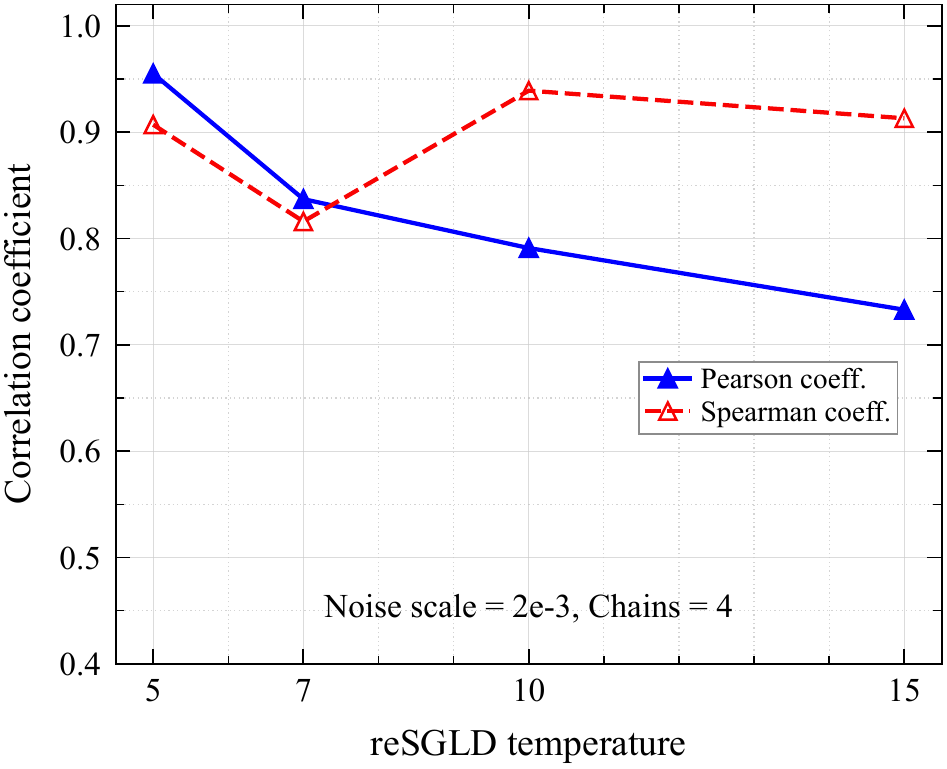}
        \caption{reSGLD temperature}
    \end{subfigure}
    % \hfill
    \begin{subfigure}[t]{0.295\linewidth}
        \centering
        \includegraphics[width=\linewidth]{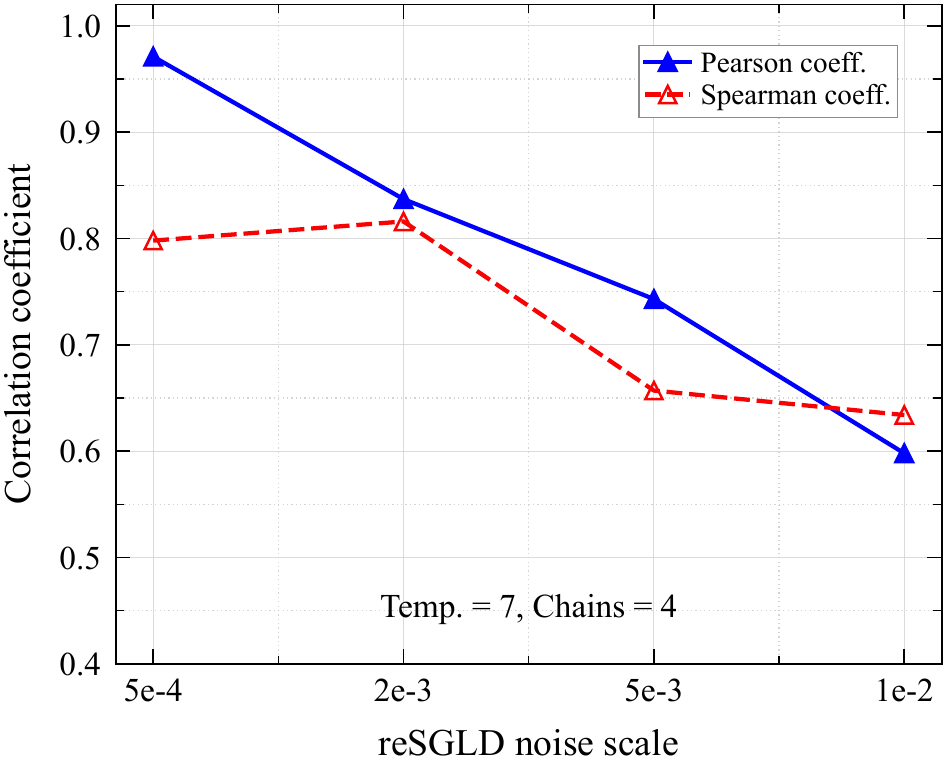}
        \caption{reSGLD noise scale}
    \end{subfigure}
    \caption{Sensitivity of the uncertainty-error correlation to the reSGLD hyperparameters in the 1D Lorenz problem. The panels show Pearson and Spearman correlation coefficients when varying the number of chains, the temperature, and the noise scale, respectively. Larger positive coefficients indicate that the posterior uncertainty more reliably tracks the true prediction error.}
    \label{fig:lorenz_correlation_sensitivity}
\end{figure}

To further assess the quality of the estimated uncertainty, we compute the Pearson and Spearman correlation coefficients between the predictive uncertainty and the corresponding relative $L_2$ error. Let $\sigma_i$ denote the average posterior standard deviation for the $i$-th trajectory and $e_i$ its relative $L_2$ error. The Pearson coefficient measures the linear correlation between $\sigma_i$ and $e_i$, whereas the Spearman coefficient evaluates their monotonic relationship based on the corresponding ranks. Larger positive coefficients indicate that trajectories with higher predictive uncertainty also tend to exhibit larger prediction errors, supporting the use of posterior uncertainty for active HF sample acquisition.

\subsection{Experiment 2: The Duffing oscillator}
\label{Sec:ex2}

The Duffing oscillator is a standard nonlinear dynamical system for modeling forced oscillations in mechanical and electrical systems. We consider the following second-order nonlinear ODE:
\begin{equation}\label{eq:duffing}
    \ddot{u} + c\dot{u} + u + u^3 = f(t),
\end{equation}
where $u(t)$ denotes the displacement, $\dot{u}(t)$ the velocity, and $\ddot{u}(t)$ the acceleration. The damping coefficient $c$ and the external forcing $f(t)$ determine the system response. Owing to the cubic nonlinearity, the Duffing oscillator provides a representative benchmark for nonlinear operator learning.

In this experiment, both the forcing amplitude and the damping coefficient are varied. The initial conditions are set to $u(0)=\dot{u}(0)=0$, and the input forcing is given by $f(t)=C\sin(2\pi t)$ for $t\in[0,20]$. The parameter pair $(C,c)$ is sampled from the two-dimensional parameter space with $C\in[6.0,8.0]$ and $c\in[0.40,0.50]$. For each parameter pair $(C,c)$, the HF trajectory is obtained by numerically solving \eqref{eq:duffing} on a fine temporal grid and taking $u(t)$ as the target output. The resulting HF inputs and outputs are discretized into 2048 time points, so that $f_H^j,u_H^j\in\mathbb{R}^{2048}$.

The LF dataset is constructed from the same family of trajectories on a coarser temporal grid with 1024 time points. Rather than using a simple affine transformation, we define the LF trajectory by a parameter-dependent non-invertible synthetic mapping
\begin{equation}
\label{eq:duffing_lf_map}
u_L(t_j;\mu)=a(\mu)\,\mathcal{D}_r\!\left[\mathcal{P}_K u_H(\alpha(\mu)t;\mu)\right]_{t=t_j}+b(t_j;\mu),
\end{equation}
where $\mu=(C,c)$, $\mathcal{P}_K$ denotes a low-pass projection,
\begin{equation}
\label{eq:duffing_lf_lowpass}
\mathcal{P}_K u_H(t;\mu)=\sum_{|k|\le K}\hat{u}_{H,k}(\mu)\exp\!\left(\frac{2\pi i k t}{T}\right),
\end{equation}
and $\mathcal{D}_r$ denotes temporal downsampling by a factor $r$. In the present implementation, we use
\begin{equation}
\label{eq:duffing_lf_params}
K=64,\qquad r=2,
\end{equation}
together with
\begin{equation}
\label{eq:duffing_lf_coeffs}
a(\mu)=0.7+0.3\,\widetilde{C},\qquad
\alpha(\mu)=0.95+0.05\,\widetilde{c},\qquad
b(t;\mu)=0.7\,\widetilde{C}\,\frac{t}{T},
\end{equation}
where $\widetilde{C},\widetilde{c}\in[-1,1]$ are the normalized parameters corresponding to $C$ and $c$, respectively. This construction removes high-frequency information through $\mathcal{P}_K$, reduces temporal resolution through $\mathcal{D}_r$, and introduces parameter-dependent amplitude and phase discrepancies through $a(\mu)$ and $\alpha(\mu)$. Accordingly, the LF inputs and outputs are represented by $f_L^j,u_L^j\in\mathbb{R}^{1024}$. In the experiments, the training candidate set consists of a $10\times10$ parameter grid, giving 100 LF/HF parameter instances in total, from which only a small number of HF samples are selected for multifidelity training.

\begin{figure}[htp!]
    \centering
    \includegraphics[width=0.8\linewidth]{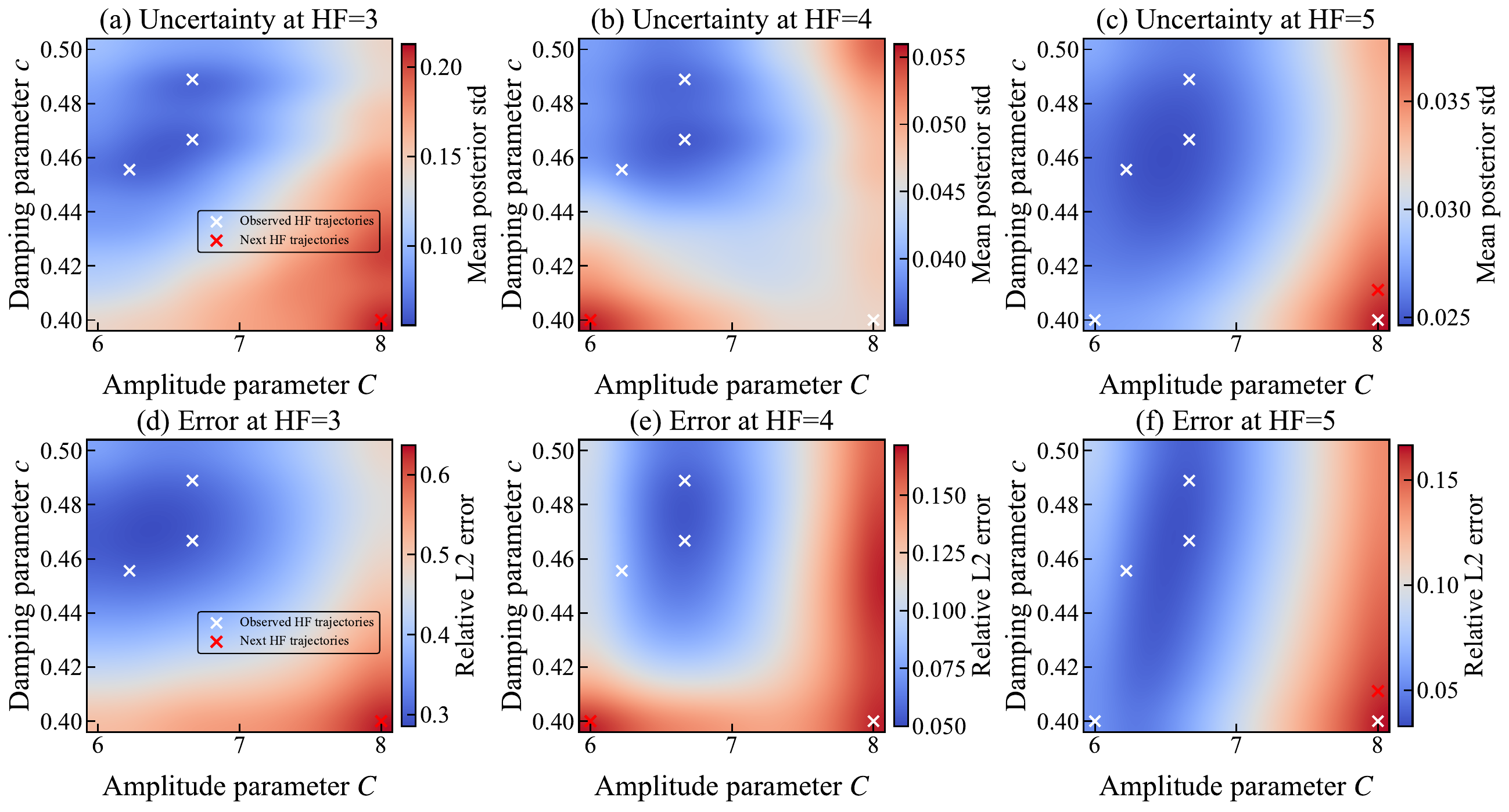}
    \caption{\textbf{Duffing oscillator:} Uncertainty-guided HF sample acquisition based on reSGLD in the $(C, c)$ parameter space. The top row shows the posterior standard deviation, and the bottom row shows the corresponding relative $L_2$ error. Red crosses indicate the selected HF samples. Regions of high posterior uncertainty largely overlap with regions of large prediction error, enabling more informative HF sample selection than random acquisition.}
    \label{fig:duffing_active_sampling}
\end{figure}

\begin{table}[htbp!]
\footnotesize
\renewcommand{\arraystretch}{1.1}
\setlength{\tabcolsep}{6pt}
\centering
\caption{Neural network specifications of the multi-fidelity operator models for the 1D Duffing problem.}
\label{tab:ex2_mf_operator_network_specs}

\begin{tabular}{lcc}
\hline
 & MF-LNO & MF-DeepONet \\
\hline

Hidden layers
&
LF: $[128]$, HF: $[2]$
&
\makecell[c]{Branch: $[256,256,256]$\\
Trunk: $[256,256,256]$}
\\

Width
&
$(3,8)$
&
$(256,256)$
\\

Activation
&
Sine
&
GELU
\\

LF parameters
&
$803$
&
$526{,}593$
\\

HF linear parameters
&
$176$
&
$1{,}444{,}097$
\\

HF nonlinear parameters
&
$176$
&
$1{,}444{,}097$
\\

\hline

Total parameters
&
$\mathbf{1{,}156}$
&
$\mathbf{3{,}414{,}788}$

\\
\hline
\end{tabular}
\end{table}

The estimated posterior uncertainty provides informative guidance for HF sample acquisition by identifying regions that are likely to exhibit large prediction errors. As shown in Fig.~\ref{fig:duffing_active_sampling}, regions of high posterior uncertainty largely overlap with those of large prediction error, supporting uncertainty-guided sample selection over random acquisition. The corresponding network architectures used for the multifidelity operator models are summarized in Table~\ref{tab:ex2_mf_operator_network_specs}.

\begin{figure}[htp!]
    \centering
    \includegraphics[width=0.8\linewidth]{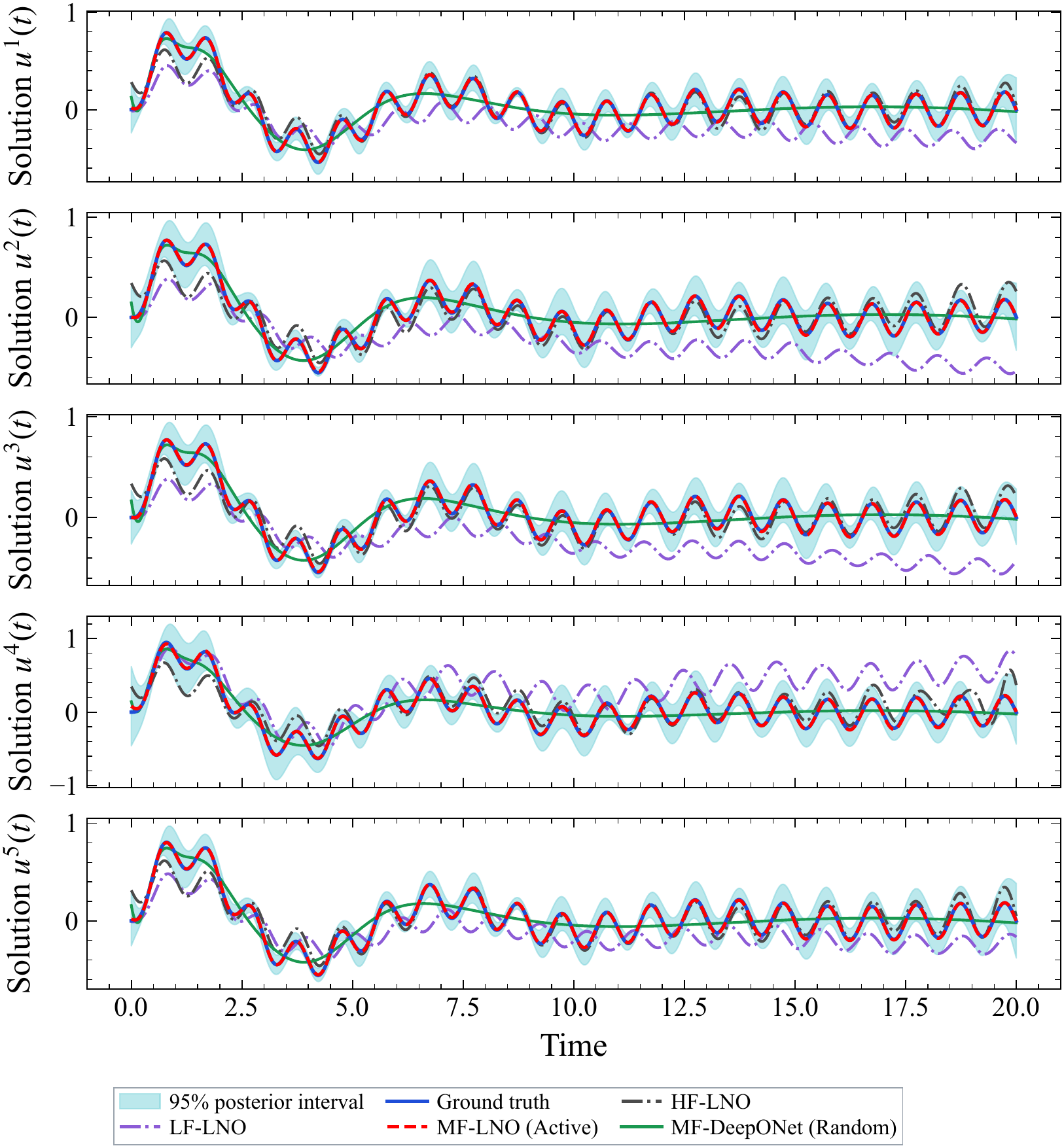}
    \caption{\textbf{1D Duffing oscillator:} Representative predictions on test samples using nine HF trajectories for multifidelity learning. The blue solid line denotes the ground truth, the red dashed line denotes MF-LNO (Active), the purple dash-dotted line denotes LF-LNO, the gray dash-dotted line denotes HF-LNO, the green solid line denotes MF-DeepONet (Random), and the cyan shaded region denotes the 95\% posterior uncertainty interval of MF-LNO (Active). MF-LNO (Active) accurately reproduces the oscillatory trajectories with only nine HF samples, whereas MF-DeepONet (Random) fails to capture the oscillatory behavior.}
    \label{fig:solution_ex2}
\end{figure}

\begin{table}[htbp!]
\centering
\footnotesize
\caption{Error comparison for the five representative test trajectories shown in Fig.~\ref{fig:solution_ex2} for the parametric Duffing oscillator. MF-LNO (Active) is trained using high-fidelity samples selected by the proposed active learning strategy, whereas MF-DeepONet (Random) is trained using randomly selected high-fidelity samples. For MF-LNO (Active), all error metrics are computed from the posterior mean prediction, while the shaded uncertainty band in Fig.~\ref{fig:solution_ex2} is constructed from the corresponding posterior standard deviation. Smaller errors are highlighted in bold for each metric within each trajectory.}
\label{tab:trajectory_error_comparison_duffing}
\begin{tabular}{llccc}
\hline
Trajectory & Model & Relative \(L^2\) error & RMSE & MAE \\
\hline
\multirow{4}{*}{\makecell[l]{Fig.~\ref{fig:solution_ex2}: Solution $u^{1}(t)$\\ ($C=6.55,\ c=0.456$)}}
& MF-LNO (Active) & \(\mathbf{2.6597\times10^{-2}}\) & \(\mathbf{6.5816\times10^{-3}}\) & \(\mathbf{5.2652\times10^{-3}}\) \\
& MF-DeepONet (Random)    & \(4.9678\times10^{-1}\)          & \(1.2293\times10^{-1}\)          & \(1.0671\times10^{-1}\) \\
& LF-LNO          & \(1.0267\times10^{0}\)          & \(2.5407\times10^{-1}\)          & \(2.0960\times10^{-1}\) \\
& HF-LNO          & \(3.0622\times10^{-1}\)          & \(7.5778\times10^{-2}\)          & \(4.9958\times10^{-2}\) \\

\hline
\multirow{4}{*}{\makecell[l]{Fig.~\ref{fig:solution_ex2}: Solution $u^{2}(t)$\\ ($C=6.30,\ c=0.418$)}}
& MF-LNO (Active) & \(\mathbf{2.9079\times10^{-2}}\) & \(\mathbf{7.1970\times10^{-3}}\) & \(\mathbf{5.8447\times10^{-3}}\) \\
& MF-DeepONet (Random)     & \(4.9309\times10^{-1}\)          & \(1.2204\times10^{-1}\)          & \(1.0462\times10^{-1}\) \\
& LF-LNO          & \(1.3628\times10^{0}\)          & \(3.3730\times10^{-1}\)          & \(2.9127\times10^{-1}\) \\
& HF-LNO          & \(4.3597\times10^{-1}\)          & \(1.0790\times10^{-1}\)          & \(8.2602\times10^{-2}\) \\

\hline
\multirow{4}{*}{\makecell[l]{Fig.~\ref{fig:solution_ex2}: Solution $u^{3}(t)$\\ ($C=6.30,\ c=0.428$)}}
& MF-LNO (Active) & \(\mathbf{2.6337\times10^{-2}}\) & \(\mathbf{6.4570\times10^{-3}}\) & \(\mathbf{5.2491\times10^{-3}}\) \\
& MF-DeepONet (Random)     & \(4.9492\times10^{-1}\)          & \(1.2134\times10^{-1}\)          & \(1.0429\times10^{-1}\) \\
& LF-LNO          & \(1.3782\times10^{0}\)          & \(3.3788\times10^{-1}\)          & \(2.9446\times10^{-1}\) \\
& HF-LNO          & \(3.9210\times10^{-1}\)          & \(9.6130\times10^{-2}\)          & \(7.2324\times10^{-2}\) \\

\hline
\multirow{4}{*}{\makecell[l]{Fig.~\ref{fig:solution_ex2}: Solution $u^{4}(t)$\\ ($C=7.87,\ c=0.442$)}}
& MF-LNO (Active) & \(\mathbf{4.5946\times10^{-2}}\) & \(\mathbf{1.3370\times10^{-2}}\) & \(\mathbf{8.6096\times10^{-3}}\) \\
& MF-DeepONet (Random)     & \(5.1446\times10^{-1}\)          & \(1.4971\times10^{-1}\)          & \(1.2942\times10^{-1}\) \\
& LF-LNO          & \(1.3729\times10^{0}\)          & \(3.9952\times10^{-1}\)          & \(3.3401\times10^{-1}\) \\
& HF-LNO          & \(4.9778\times10^{-1}\)          & \(1.4485\times10^{-1}\)          & \(1.0774\times10^{-1}\) \\

\hline
\multirow{4}{*}{\makecell[l]{Fig.~\ref{fig:solution_ex2}: Solution $u^{5}(t)$\\ ($C=6.64,\ c=0.445$)}}
& MF-LNO (Active) & \(\mathbf{2.6971\times10^{-2}}\) & \(\mathbf{6.8130\times10^{-3}}\) & \(\mathbf{5.3487\times10^{-3}}\) \\
& MF-DeepONet (Random)     & \(4.9626\times10^{-1}\)          & \(1.2536\times10^{-1}\)          & \(1.0853\times10^{-1}\) \\
& LF-LNO          & \(8.8228\times10^{-1}\)          & \(2.2287\times10^{-1}\)          & \(1.8984\times10^{-1}\) \\
& HF-LNO          & \(3.4430\times10^{-1}\)          & \(8.6973\times10^{-2}\)          & \(5.8889\times10^{-2}\) \\

\hline
\end{tabular}
\end{table}

\begin{figure}[htp!]
    \centering
    \begin{subfigure}[t]{0.49\linewidth}
        \centering
        \includegraphics[width=\linewidth]{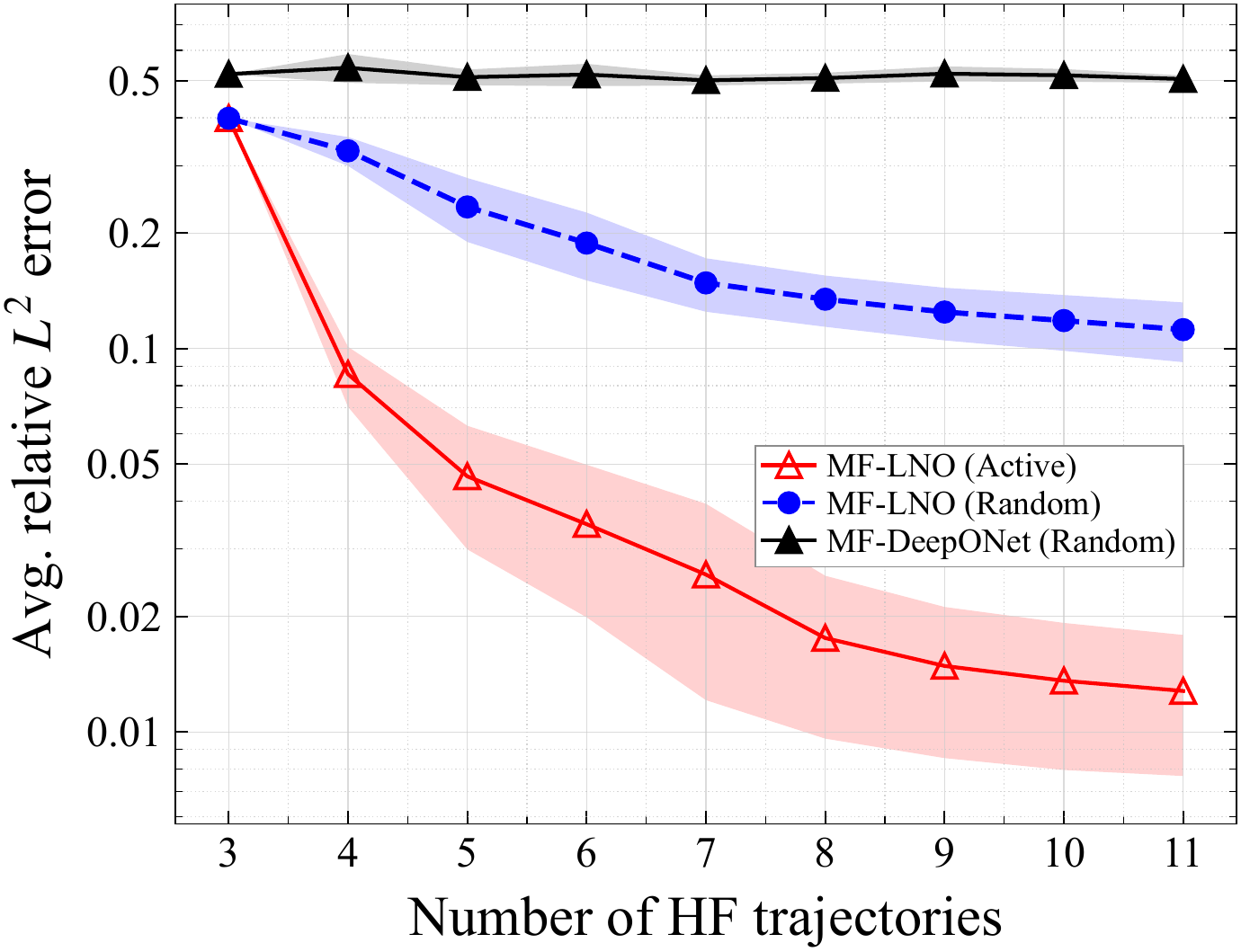}
        \caption{Avg. $L^2$ relative error}
    \end{subfigure}
    \hfill
    \begin{subfigure}[t]{0.49\linewidth}
        \centering
        \includegraphics[width=\linewidth]{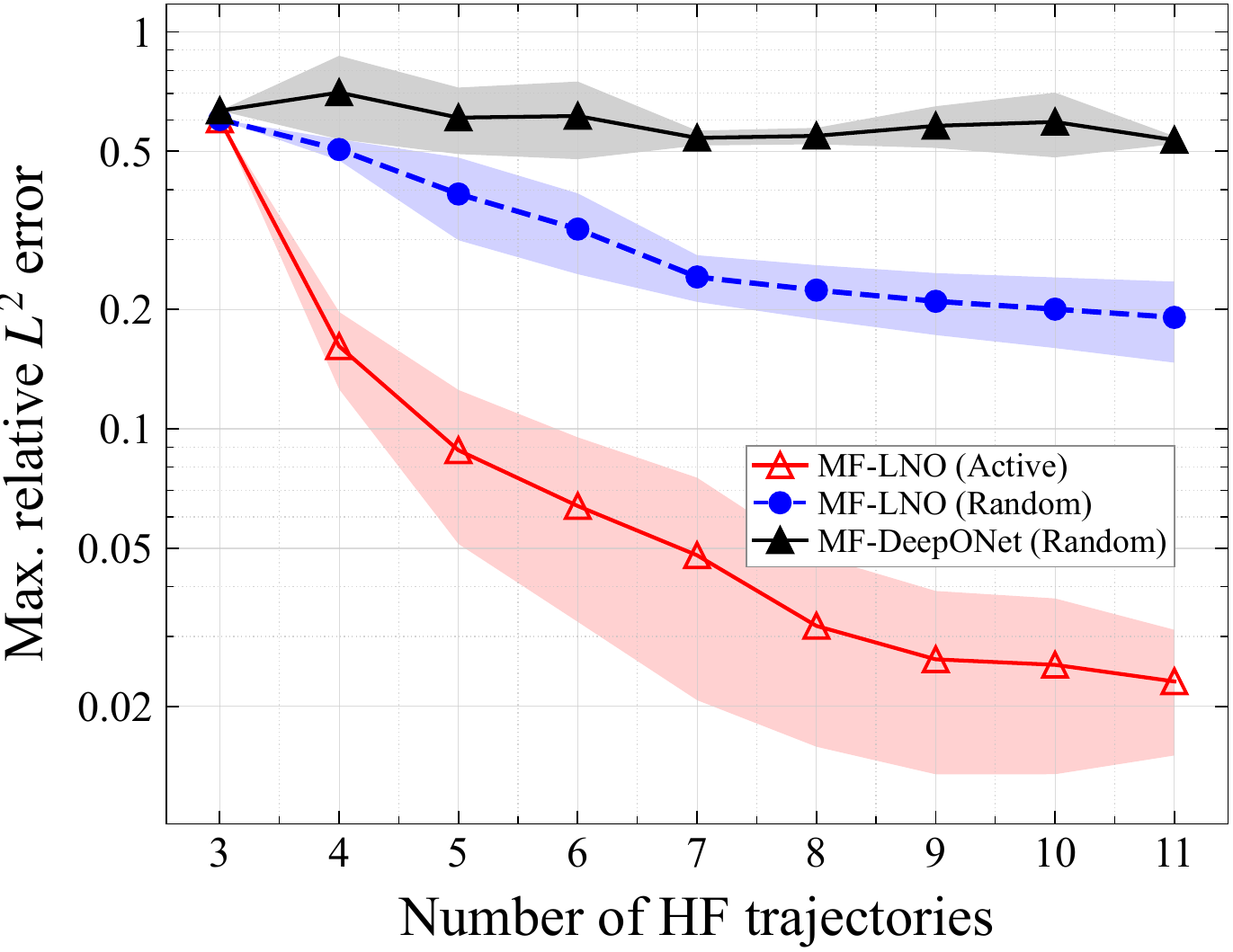}
        \caption{Max. $L^2$ relative error}
    \end{subfigure}
    \caption{\textbf{1D Duffing oscillator:} Error comparison across five random seeds as the number of HF trajectories increases. The left panel shows the average relative \(L_2\) error, and the right panel shows the maximum relative \(L_2\) error. MF-LNO (Active) employs uncertainty-guided HF trajectory acquisition, whereas MF-LNO (Random) and MF-DeepONet (Random) use random HF trajectory acquisition. All methods are initialized with the same three HF trajectories, and the shaded bands indicate the standard error over the five seeds.}
    \label{fig:duffing_error_comparison}
\end{figure}

\begin{table}[htbp!]
\centering
\footnotesize
\caption{Comparison of the average and maximum relative \(L^2\) errors over all test trajectories for the parametric Duffing oscillator with different numbers of high-fidelity (HF) trajectories. MF-LNO (Active) acquires HF trajectories using the proposed active learning strategy based on posterior uncertainty, whereas MF-LNO (Random) and MF-DeepONet (Random) employ random HF trajectory acquisition. All methods are initialized with the same three HF trajectories. The smallest error in each row is highlighted in bold.}
\label{tab:duffing_error_comparison_compact}
\begin{tabular}{clccc}
\hline
\makecell[c]{Number of HF \\ trajectories}
& Metric
& \makecell[c]{MF-LNO \\ (Active)}
& \makecell[c]{MF-LNO \\ (Random)}
& \makecell[c]{MF-DeepONet \\ (Random)} \\
\hline
\multirow{2}{*}{3}
& Avg. error
& \(\mathbf{3.9877\times10^{-1}}\)
& \(\mathbf{3.9877\times10^{-1}}\)
& \(5.1995\times10^{-1}\) \\
& Max error
& \(\mathbf{6.0123\times10^{-1}}\)
& \(\mathbf{6.0123\times10^{-1}}\)
& \(6.3303\times10^{-1}\) \\
\hline
\multirow{2}{*}{5}
& Avg. error
& \(\mathbf{7.5590\times10^{-2}}\)
& \(2.3598\times10^{-1}\)
& \(4.8805\times10^{-1}\) \\
& Max error
& \(\mathbf{1.5818\times10^{-1}}\)
& \(4.3391\times10^{-1}\)
& \(5.2570\times10^{-1}\) \\
\hline
\multirow{2}{*}{7}
& Avg. error
& \(\mathbf{5.1931\times10^{-2}}\)
& \(1.2681\times10^{-1}\)
& \(4.9294\times10^{-1}\) \\
& Max error
& \(\mathbf{1.0192\times10^{-1}}\)
& \(2.1105\times10^{-1}\)
& \(5.2948\times10^{-1}\) \\
\hline
\multirow{2}{*}{9}
& Avg. error
& \(\mathbf{2.7403\times10^{-2}}\)
& \(9.9472\times10^{-2}\)
& \(5.3890\times10^{-1}\) \\
& Max error
& \(\mathbf{5.0325\times10^{-2}}\)
& \(1.6124\times10^{-1}\)
& \(5.6925\times10^{-1}\) \\
\hline
\end{tabular}
\end{table}

Using only nine HF trajectories, the proposed Bayesian MF-LNO accurately captures the nonlinear oscillatory dynamics of the Duffing system while providing well-calibrated predictive uncertainty. As shown in Fig.~\ref{fig:solution_ex2}, the proposed method closely matches the ground-truth trajectories, whereas the HF-LNO, LF-LNO, and MF-DeepONet exhibit noticeably larger prediction errors. This qualitative observation is quantitatively confirmed in Table~\ref{tab:trajectory_error_comparison_duffing}, where the proposed method consistently achieves the lowest relative $L_2$ error, RMSE, and MAE across all representative test trajectories. As additional HF trajectories are acquired, the prediction errors decrease consistently, with the proposed uncertainty-guided acquisition achieving substantially faster error reduction than random acquisition, as shown in Fig.~\ref{fig:duffing_error_comparison}. The same trend is reflected in Table~\ref{tab:duffing_error_comparison_compact}, where the proposed Bayesian MF-LNO consistently attains the lowest average and maximum relative $L_2$ errors for all HF sample sizes.

% To learn the mapping from $f(t)$ to $u(t)$, we train the LF component of MF-LNO using a 5-chain reSGLD sampler for 2,000 epochs to map $f_L(t)$ to $u_L(t)$. The trained LNO in Phase 1 is then refined with HF data, training the MF-LNO for 5,000 epochs with another 5-chain reSGLD sampler, with 10 ensemble models sampled every 100 epochs. The training batch sizes are 20 and 10 respectively. Fig. \ref{fig:duffing1} shows the predictive means and 95\% confidence intervals for six randomly chosen input functions with different amplitudes $A$. Comparative results with LF predictions, HF predictions, and Mix predictions are presented in Fig. \ref{fig:duffing2}. From the result, it is clear that the Mix predictions and the HF predictions lead to trivial results. This behavior may stem from poor initialization and the difficulty of the sampler extracting patterns from the limited HF data. LF predictions, while avoiding underfitting, still suffer from bias according to the approximated output functions.

\subsection{Experiment 3: Composite beam}
\label{Sec:ex3}

\begin{figure}[htp!]
    \centering
    \includegraphics[width=1.0\linewidth]{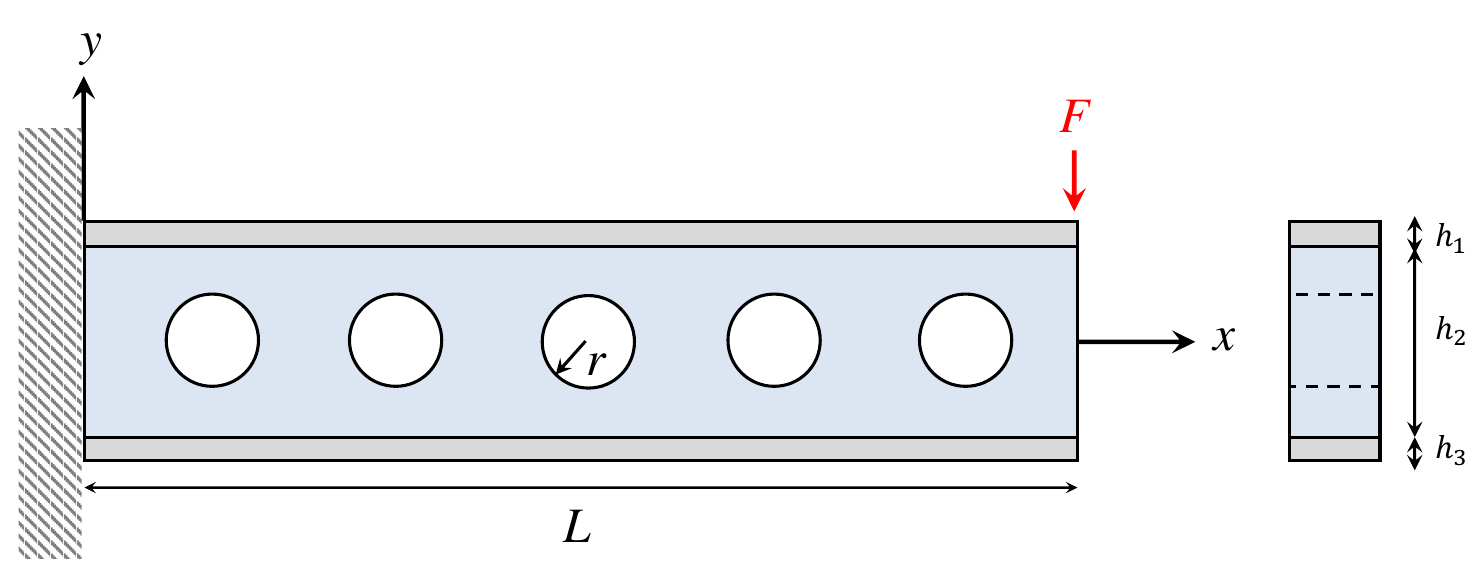}
    \caption{\textbf{Composite beam.} The high-fidelity model is a two-dimensional plane-stress perforated composite beam, in which the left boundary is fixed and a downward traction is prescribed on the right boundary. The beam has length $L$, the circular holes have radius $r$, and the layered cross-section consists of top, core, and bottom parts with thicknesses $h_1$, $h_2$, and $h_3$, respectively.}
    \label{fig:ex3_configuration}
\end{figure}

We next consider a more realistic multifidelity beam dynamics problem, in which the discrepancy between the two fidelity levels is not limited to simple scale differences or response bias, but also includes differences in oscillation frequency induced by the change of model fidelity. As illustrated in Fig.~\ref{fig:ex3_configuration}, the problem consists of a HF two-dimensional perforated composite cantilever beam and a LF one-dimensional Euler--Bernoulli beam. The beam geometry is fixed, and the parameter space is taken as $(a_1,\xi_2)$, where $a_1$ denotes the load amplitude and $\xi_2$ denotes the Young's modulus of the core layer.

\begin{figure}[htp!]
    \centering
    \includegraphics[width=1.0\linewidth]{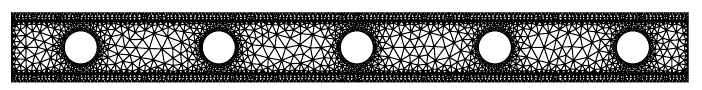}
    \caption{HF mesh for the composite beam. Finite element mesh of the two-dimensional perforated composite beam used in the high-fidelity model.}
    \label{fig:ex3_mesh}
\end{figure}

For the HF model, which is discretized using the finite element mesh shown in Fig.~\ref{fig:ex3_mesh}, letting $\bm{u}(\bm{x},t)=[u_x(\bm{x},t),u_y(\bm{x},t)]^\top$, we solve
\begin{equation}
\rho(\bm{x})\,\bm{u}_{tt}+c_H\bm{u}_t-\nabla\cdot\bm{\sigma}(\bm{u})=\bm{0},
\qquad \bm{x}\in\Omega,\quad t\in(0,T],
\end{equation}
with $\bm{u}=\bm{0}$ on the left boundary and a downward traction prescribed on the right boundary. Here $\bm{\sigma}(\bm{u})=\bm{C}(\bm{x};\xi_1,\xi_2,\xi_3):\bm{\varepsilon}(\bm{u})$ and $\bm{\varepsilon}(\bm{u})=\frac12(\nabla\bm{u}+\nabla\bm{u}^\top)$. The HF model is solved by a two-dimensional finite element discretization together with Newmark time integration.

\begin{figure}[htp!]
    \centering
    \begin{subfigure}[t]{1\linewidth}
        \centering
        \includegraphics[width=\linewidth]{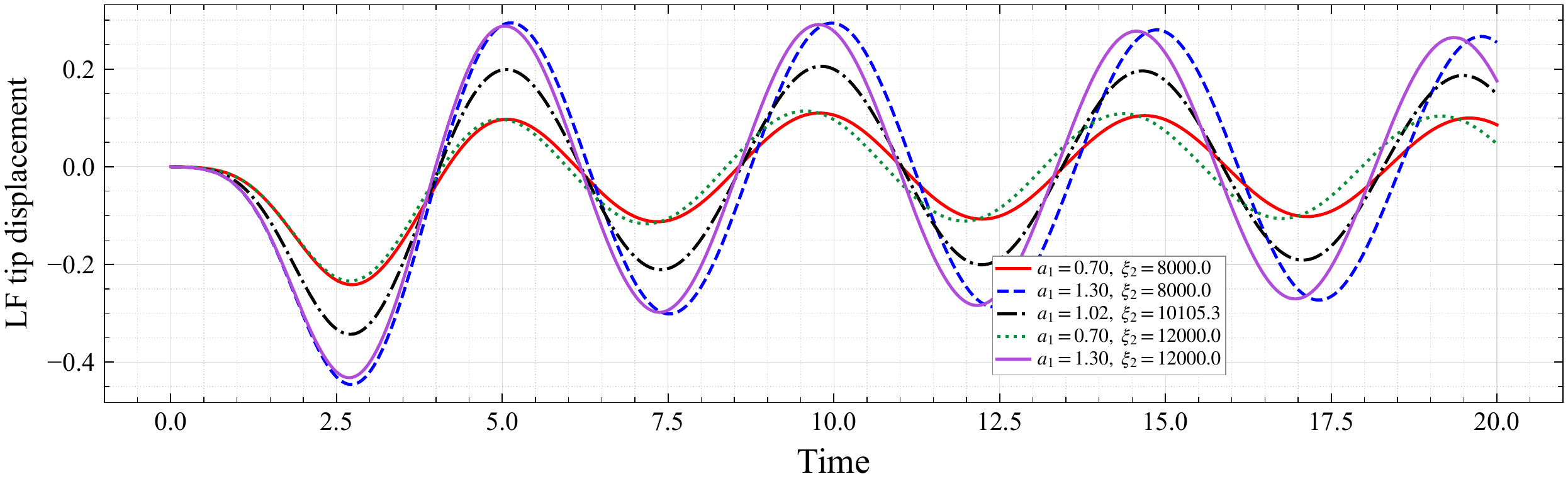}
        \caption{LF tip trajectories}
    \end{subfigure}
    \hfill \\
    \begin{subfigure}[t]{1\linewidth}
        \centering
        \includegraphics[width=\linewidth]{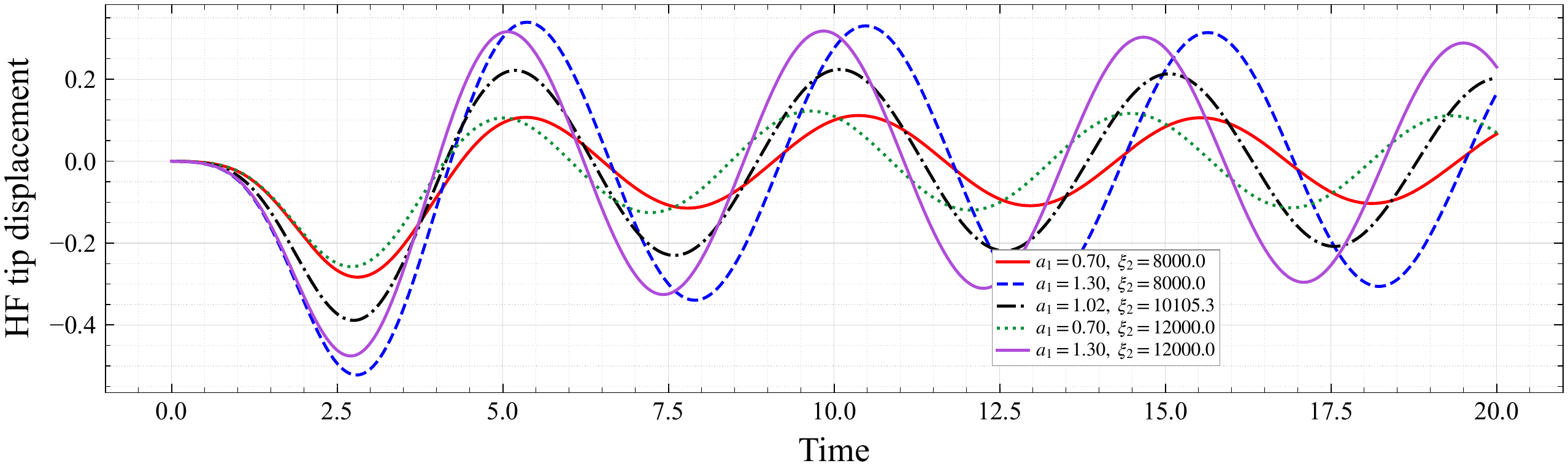}
        \caption{HF tip trajectories}
    \end{subfigure}
    \caption{Representative LF and HF tip-displacement trajectories for the composite beam problem over the sampled parameter space $(a_1,\xi_2)$. The top panel shows the LF responses obtained from the one-dimensional Euler--Bernoulli beam model, and the bottom panel shows the corresponding HF responses obtained from the two-dimensional perforated composite beam model. The two fidelity levels exhibit differences in oscillation frequency.}
    \label{fig:beam_lf_hf_tip_trajectories}
\end{figure}

\begin{table}[htbp!]
\footnotesize
\renewcommand{\arraystretch}{1.1}
\setlength{\tabcolsep}{6pt}
\centering
\caption{Neural network specifications of the multi-fidelity operator models for the composite beam problem.}
\label{tab:ex3_mf_operator_network_specs}

\begin{tabular}{lcc}
\hline
 & MF-LNO & MF-DeepONet \\
\hline

Hidden layers
&
LF: $[128]$, HF: $[2]$
&
\makecell[c]{Branch: $[32,32,32]$\\
Trunk: $[32,32,32]$}
\\

Width
&
$(3,8)$
&
$(32,32)$
\\

Activation
&
Sine
&
GELU
\\

LF parameters
&
$806$
&
$9{,}697$
\\

HF linear parameters
&
$179$
&
$12{,}929$
\\

HF nonlinear parameters
&
$179$
&
$12{,}929$
\\

\hline

Total parameters
&
$\mathbf{1{,}165}$
&
$\mathbf{35{,}556}$

\\
\hline
\end{tabular}
\end{table}

For the LF model, letting $w(x,t)$ denote the transverse displacement, we solve
\begin{equation}
\rho_L w_{tt}+c_L w_t+\kappa_L w_{xxxx}=0,
\qquad x\in(0,L),\quad t\in(0,T],
\end{equation}
with boundary conditions
\begin{equation}
w(0,t)=0,\qquad w_x(0,t)=0,\qquad w_{xx}(L,t)=0,\qquad
\kappa_L w_{xxx}(L,t)=g(t).
\end{equation}
Here $\kappa_L$ denotes the effective bending stiffness of the LF beam and is taken as a simplified function of $\xi_2$. The LF model is solved by a one-dimensional beam finite difference discretization.

Both models start from rest and are driven by the same time-dependent tip load. We use a Gaussian-mixture load $g(t)=a_1\exp\!\bigl(-(t-\tau_1)^2/(2\sigma_1^2)\bigr)+a_2\exp\!\bigl(-(t-\tau_2)^2/(2\sigma_2^2)\bigr)$, but only $a_1$ is varied in the experiments; the remaining load-shape parameters are fixed, for example with $a_2=0.35$, $\tau_1=1.5$, $\tau_2=4.2$, $\sigma_1=0.55$, and $\sigma_2=0.8$. Samples are generated over $(a_1,\xi_2)$ with $a_1\in[0.7,1.3]$ and $\xi_2\in[8000,12000]$ on a uniform $20\times20$ grid. The resulting response operators are $\mathcal{G}_H:(a_1,\xi_2)\mapsto u_y^{\mathrm{tip}}(t)$ and $\mathcal{G}_L:(a_1,\xi_2)\mapsto w^{\mathrm{tip}}(t)$.

The discrepancy between the LF and HF models is illustrated in Fig.~\ref{fig:beam_lf_hf_tip_trajectories}. Unlike the previous benchmarks, the two fidelity levels exhibit noticeable differences in oscillation frequency, making the multifidelity mapping substantially more challenging than simple amplitude or bias correction. The network architectures used for the composite beam problem are summarized in Table~\ref{tab:ex3_mf_operator_network_specs}.

\begin{figure}[htp!]
    \centering
    \includegraphics[width=1.0\linewidth]{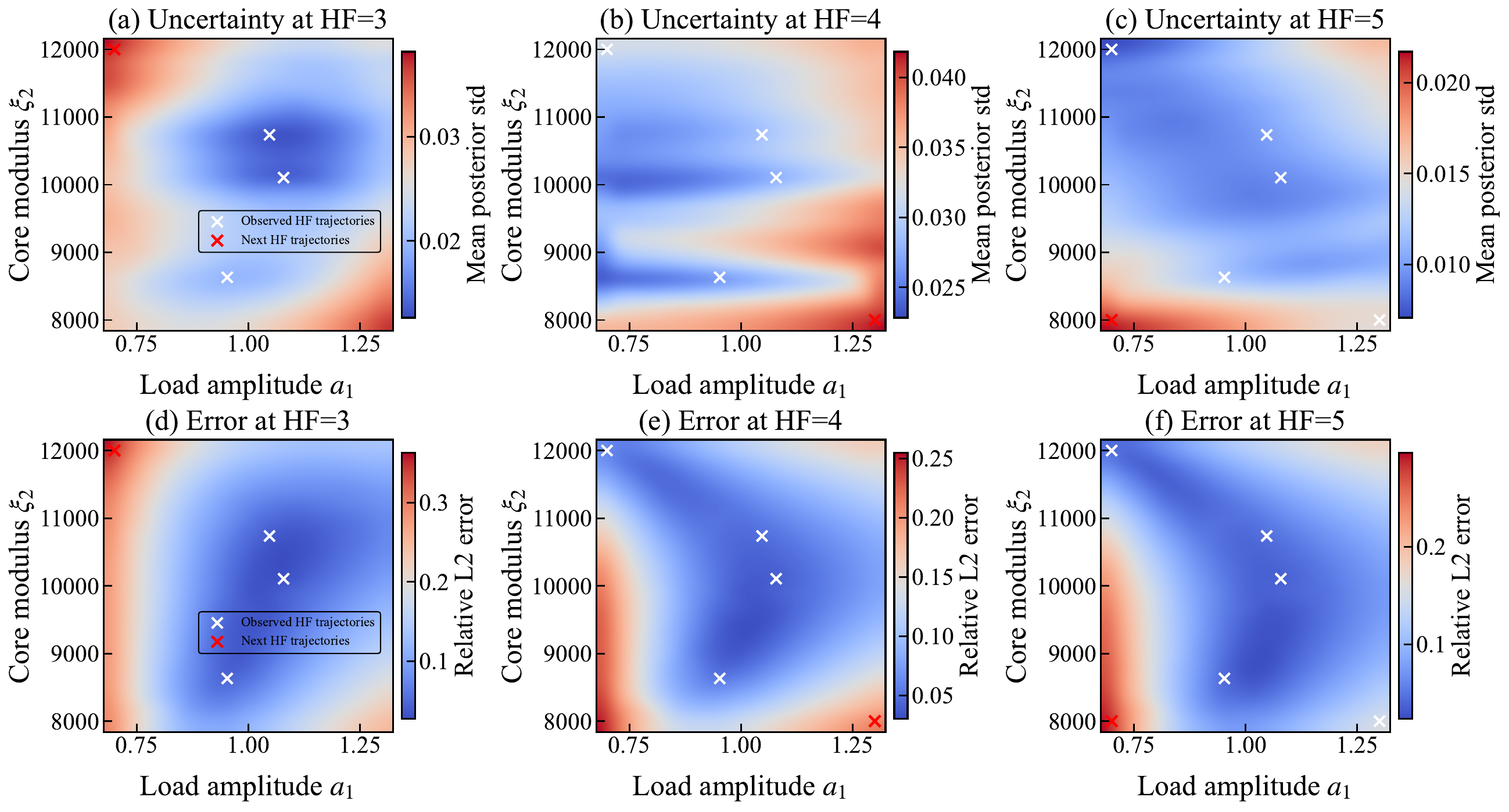}
    \caption{Uncertainty-guided sample acquisition based on reSGLD in the $(a_1,\xi_2)$ parameter space. The top row shows the posterior standard deviation, and the bottom row shows the corresponding relative $L_2$ error. Red crosses denote the selected samples. The overlap between regions of large uncertainty and prediction error supports uncertainty-guided sample acquisition.}
    \label{fig:beam_active_sampling}
\end{figure}

\begin{figure}[htp!]
    \centering
    \includegraphics[width=1.0\linewidth]{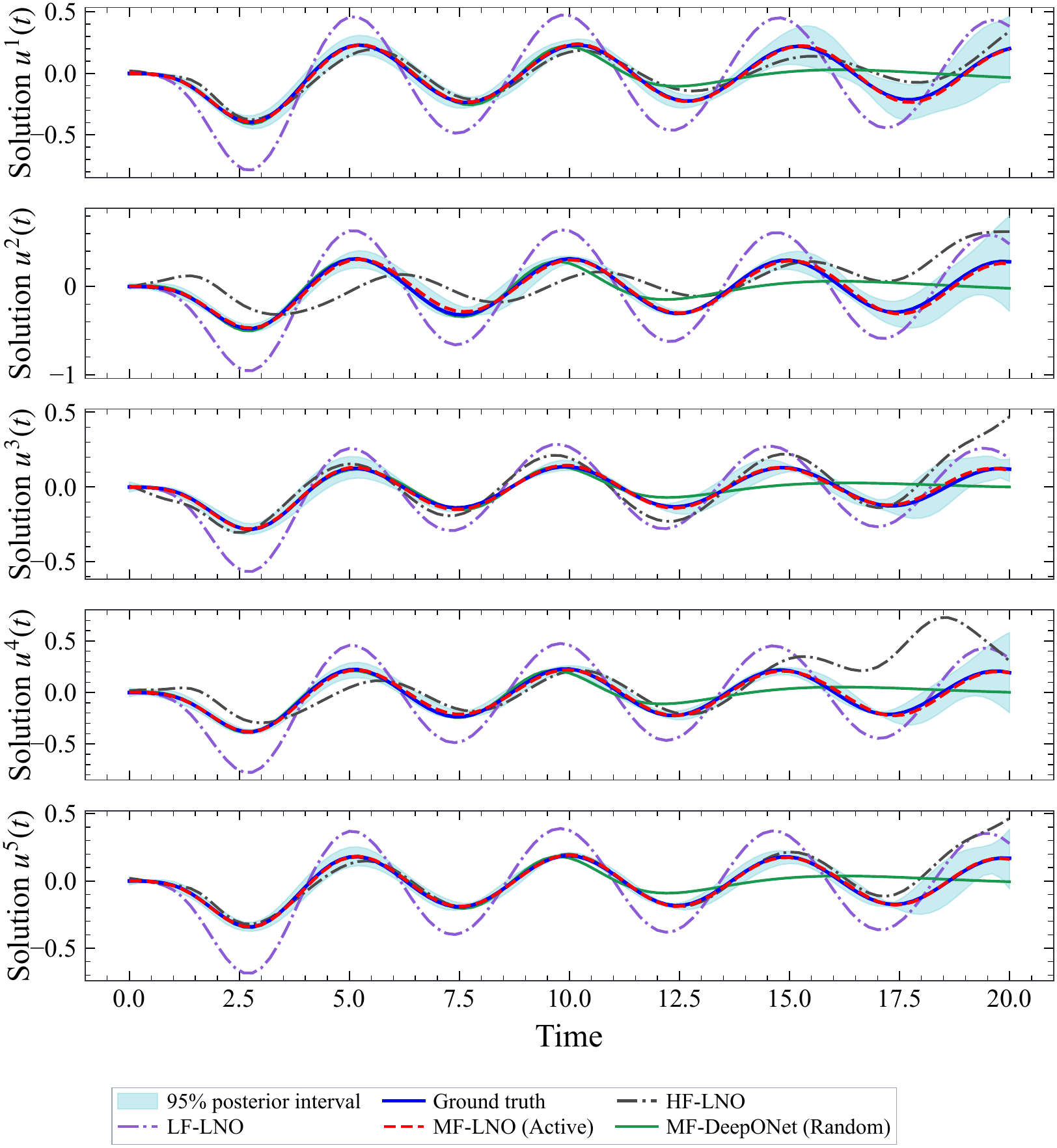}
    \caption{\textbf{Composite beam:} Representative predictions on test samples using eleven HF trajectories for multifidelity learning. The blue solid line denotes the ground truth, the red dashed line denotes MF-LNO (Active), the purple dash-dotted line denotes LF-LNO, the gray dash-dotted line denotes HF-LNO, the green solid line denotes MF-DeepONet (Random), and the cyan shaded region denotes the 95\% posterior uncertainty interval of MF-LNO (Active). MF-LNO (Active) accurately reproduces the tip-displacement trajectories with only eleven HF samples, whereas MF-DeepONet (Random) yields substantially less accurate predictions.}
    \label{fig:solution_ex3}
\end{figure}

\begin{table}[htbp!]
\centering
\footnotesize
\caption{Error comparison for the five representative test trajectories shown in Fig.~\ref{fig:solution_ex3} for the 2D composite beam problem. MF-LNO (Active) is trained using high-fidelity samples selected by the proposed active learning strategy, whereas MF-DeepONet (Random) is trained using randomly selected high-fidelity samples. For MF-LNO (Active), all error metrics are computed from the posterior mean prediction, while the shaded uncertainty band in Fig.~\ref{fig:solution_ex3} is constructed from the corresponding posterior standard deviation. Smaller errors are highlighted in bold for each metric within each trajectory.}
\label{tab:trajectory_error_comparison_beam}
\begin{tabular}{llccc}
\hline
Trajectory & Model & Relative \(L^2\) error & RMSE & MAE \\
\hline
\multirow{4}{*}{\makecell[l]{Fig.~\ref{fig:solution_ex3}: Solution $u^{1}(t)$\\ ($a_1=1.032,\ \xi_2=9894.7$)}}
& MF-LNO (Active) & \(\mathbf{5.1014\times10^{-2}}\) & \(\mathbf{9.0049\times10^{-3}}\) & \(\mathbf{7.0275\times10^{-3}}\) \\
& MF-DeepONet (Random)     & \(5.1957\times10^{-1}\)          & \(9.1713\times10^{-2}\)          & \(5.9338\times10^{-2}\) \\
& LF-LNO          & \(5.6327\times10^{-1}\)          & \(9.9427\times10^{-2}\)          & \(8.5299\times10^{-2}\) \\
& HF-LNO          & \(3.7185\times10^{-1}\)          & \(6.5637\times10^{-2}\)          & \(5.4666\times10^{-2}\) \\

\hline
\multirow{4}{*}{\makecell[l]{Fig.~\ref{fig:solution_ex3}: Solution $u^{2}(t)$\\ ($a_1=1.280,\ \xi_2=11157.9$)}}
& MF-LNO (Active) & \(\mathbf{1.0076\times10^{-1}}\) & \(\mathbf{2.3334\times10^{-2}}\) & \(\mathbf{1.9192\times10^{-2}}\) \\
& MF-DeepONet (Random)     & \(5.7001\times10^{-1}\)          & \(1.3200\times10^{-1}\)          & \(8.7968\times10^{-2}\) \\
& LF-LNO          & \(4.2923\times10^{-1}\)          & \(9.9402\times10^{-2}\)          & \(8.6799\times10^{-2}\) \\
& HF-LNO          & \(1.0347\times10^{0}\)          & \(2.3962\times10^{-1}\)          & \(2.0445\times10^{-1}\) \\

\hline
\multirow{4}{*}{\makecell[l]{Fig.~\ref{fig:solution_ex3}: Solution $u^{3}(t)$\\ ($a_1=0.752,\ \xi_2=10947.4$)}}
& MF-LNO (Active) & \(\mathbf{6.8705\times10^{-2}}\) & \(\mathbf{7.7141\times10^{-3}}\) & \(\mathbf{5.9543\times10^{-3}}\) \\
& MF-DeepONet (Random)     & \(5.0871\times10^{-1}\)          & \(5.7117\times10^{-2}\)          & \(3.6989\times10^{-2}\) \\
& LF-LNO          & \(4.5517\times10^{-1}\)          & \(5.1106\times10^{-2}\)          & \(4.4181\times10^{-2}\) \\
& HF-LNO          & \(8.3361\times10^{-1}\)          & \(9.3596\times10^{-2}\)          & \(6.7522\times10^{-2}\) \\

\hline
\multirow{4}{*}{\makecell[l]{Fig.~\ref{fig:solution_ex3}: Solution $u^{4}(t)$\\ ($a_1=1.032,\ \xi_2=11578.9$)}}
& MF-LNO (Active) & \(\mathbf{8.5752\times10^{-2}}\) & \(\mathbf{1.4916\times10^{-2}}\) & \(\mathbf{1.1980\times10^{-2}}\) \\
& MF-DeepONet (Random)     & \(5.6088\times10^{-1}\)          & \(9.7558\times10^{-2}\)          & \(6.3671\times10^{-2}\) \\
& LF-LNO          & \(4.1170\times10^{-1}\)          & \(7.1611\times10^{-2}\)          & \(6.2363\times10^{-2}\) \\
& HF-LNO          & \(1.4361\times10^{0}\)          & \(2.4980\times10^{-1}\)          & \(1.6384\times10^{-1}\) \\

\hline
\multirow{4}{*}{\makecell[l]{Fig.~\ref{fig:solution_ex3}: Solution $u^{5}(t)$\\ ($a_1=0.908,\ \xi_2=10947.4$)}}
& MF-LNO (Active) & \(\mathbf{3.0803\times10^{-2}}\) & \(\mathbf{4.5193\times10^{-3}}\) & \(\mathbf{3.7581\times10^{-3}}\) \\
& MF-DeepONet (Random)     & \(5.3376\times10^{-1}\)          & \(7.8314\times10^{-2}\)          & \(5.0182\times10^{-2}\) \\
& LF-LNO          & \(4.5599\times10^{-1}\)          & \(6.6903\times10^{-2}\)          & \(5.8100\times10^{-2}\) \\
& HF-LNO          & \(5.3604\times10^{-1}\)          & \(7.8648\times10^{-2}\)          & \(4.4264\times10^{-2}\) \\

\hline
\end{tabular}
\end{table}

\begin{figure}[htp!]
    \centering
    \begin{subfigure}[t]{0.49\linewidth}
        \centering
        \includegraphics[width=\linewidth]{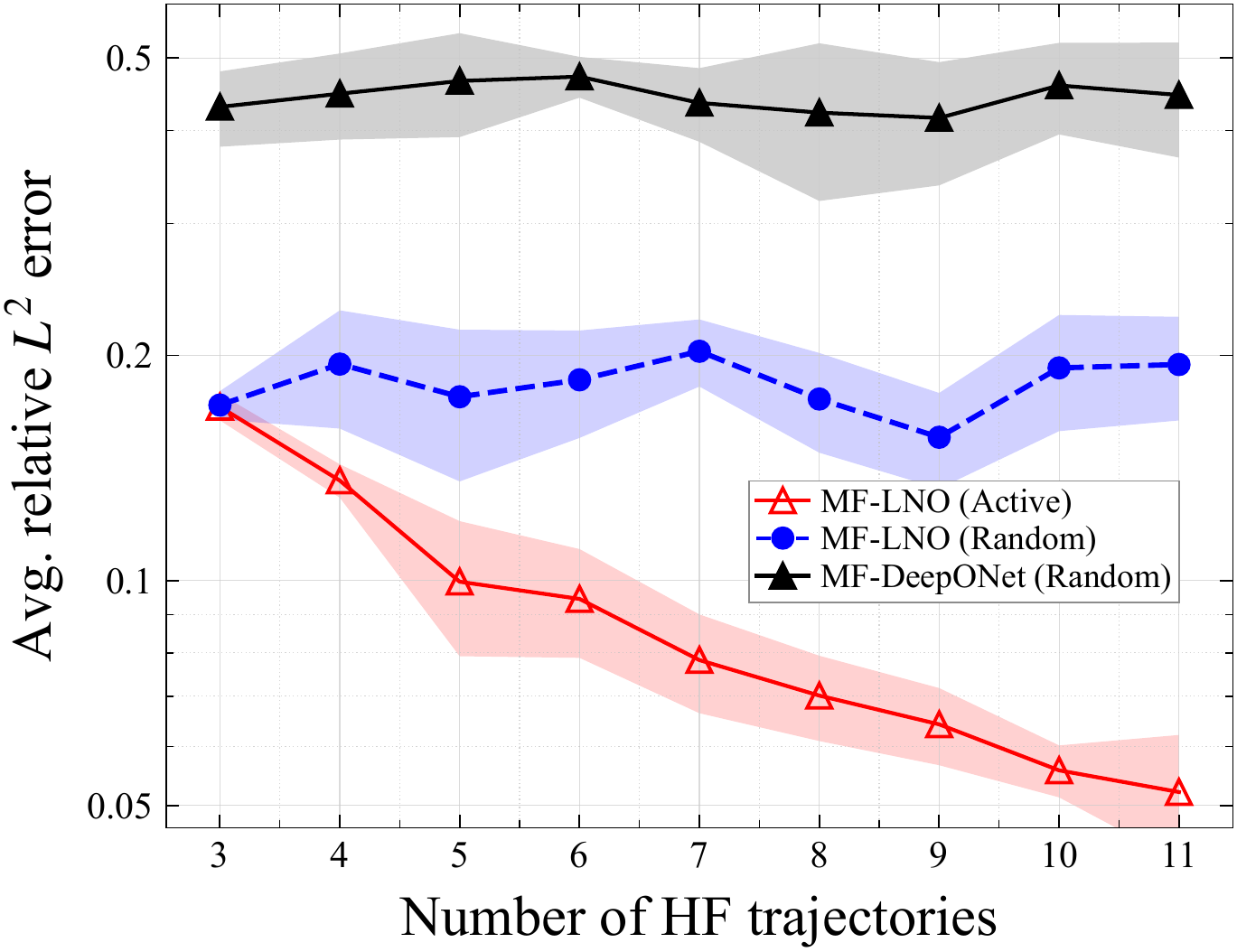}
        \caption{Avg. $L^2$ relative error}
    \end{subfigure}
    \hfill
    \begin{subfigure}[t]{0.49\linewidth}
        \centering
        \includegraphics[width=\linewidth]{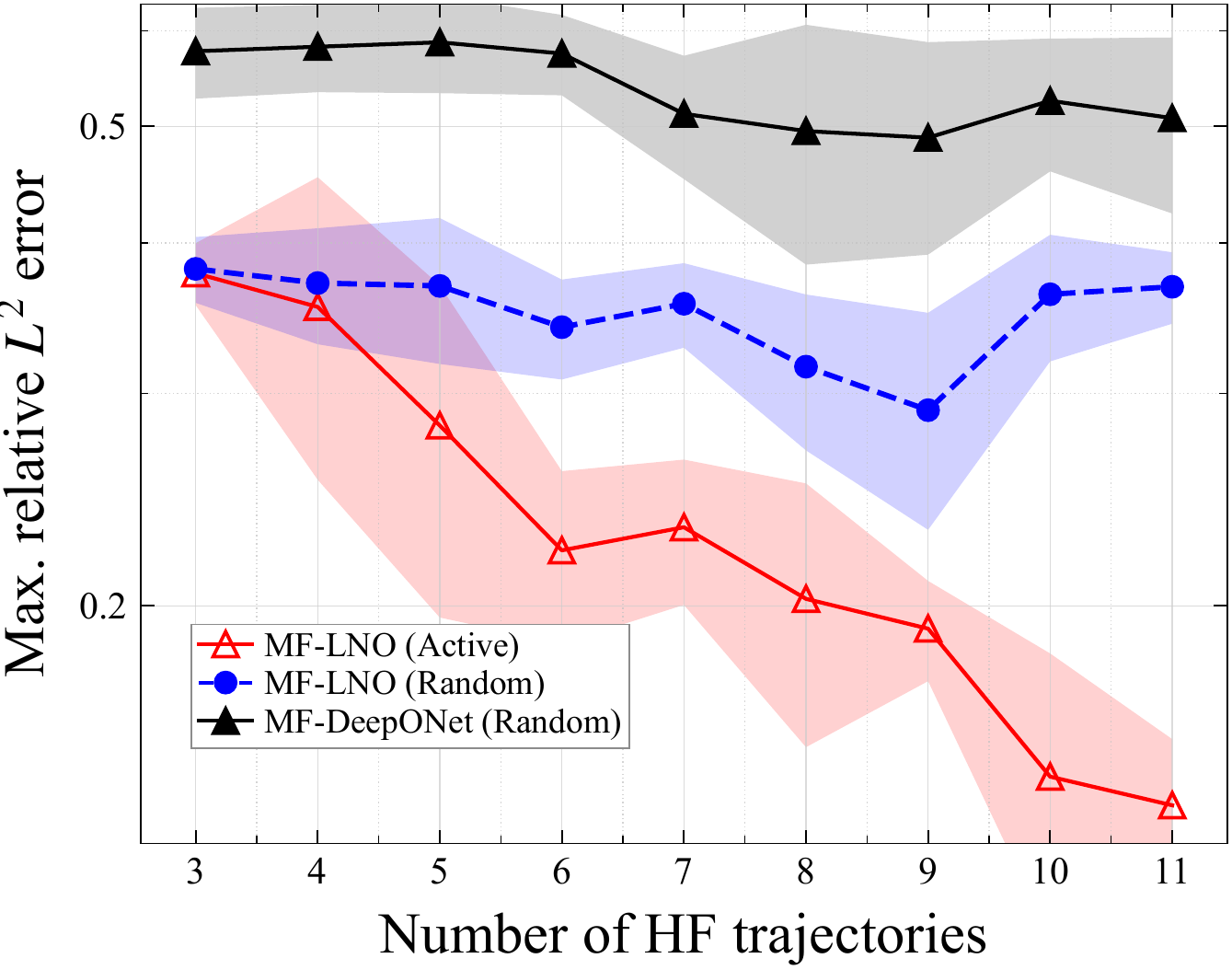}
        \caption{Max. $L^2$ relative error}
    \end{subfigure}
    \caption{\textbf{Composite beam:} Error comparison across five random seeds as the number of HF trajectories increases. The left panel shows the average relative \(L_2\) error, and the right panel shows the maximum relative \(L_2\) error. MF-LNO (Active) employs uncertainty-guided HF trajectory acquisition, whereas MF-LNO (Random) and MF-DeepONet (Random) use random HF trajectory acquisition. All methods are initialized with the same three HF trajectories, and the shaded bands indicate the standard deviation over the five seeds.}
    \label{fig:beam_error_comparison}
\end{figure}

\begin{table}[htbp!]
\centering
\footnotesize
\caption{Comparison of the average and maximum relative \(L^2\) errors over all test trajectories for the composite beam problem with different numbers of high-fidelity (HF) trajectories. MF-LNO (Active) acquires HF trajectories using the proposed active learning strategy based on posterior uncertainty, whereas MF-LNO (Random) and MF-DeepONet (Random) employ random HF trajectory acquisition. All methods are initialized with the same three HF trajectories. The smallest error in each row is highlighted in bold.}
\label{tab:beam_error_comparison_compact}
\begin{tabular}{clccc}
\hline
\makecell[c]{Number of HF \\ trajectories}
& Metric
& \makecell[c]{MF-LNO \\ (Active)}
& \makecell[c]{MF-LNO \\ (Random)}
& \makecell[c]{MF-DeepONet \\ (Random)} \\
\hline
\multirow{2}{*}{3}
& Avg. error
& \(\mathbf{1.6699\times10^{-1}}\)
& \(\mathbf{1.6699\times10^{-1}}\)
& \(4.2047\times10^{-1}\) \\
& Max error
& \(\mathbf{3.6495\times10^{-1}}\)
& \(\mathbf{3.6495\times10^{-1}}\)
& \(5.4307\times10^{-1}\) \\
\hline
\multirow{2}{*}{5}
& Avg. error
& \(\mathbf{9.7926\times10^{-2}}\)
& \(2.1622\times10^{-1}\)
& \(4.9198\times10^{-1}\) \\
& Max error
& \(\mathbf{3.4268\times10^{-1}}\)
& \(3.9837\times10^{-1}\)
& \(5.6807\times10^{-1}\) \\
\hline
\multirow{2}{*}{7}
& Avg. error
& \(\mathbf{7.4235\times10^{-2}}\)
& \(2.1686\times10^{-1}\)
& \(4.2598\times10^{-1}\) \\
& Max error
& \(\mathbf{2.3340\times10^{-1}}\)
& \(3.6402\times10^{-1}\)
& \(5.2802\times10^{-1}\) \\
\hline
\multirow{2}{*}{9}
& Avg. error
& \(\mathbf{6.9349\times10^{-2}}\)
& \(1.2275\times10^{-1}\)
& \(5.1344\times10^{-1}\) \\
& Max error
& \(\mathbf{2.0569\times10^{-1}}\)
& \(2.3091\times10^{-1}\)
& \(5.8806\times10^{-1}\) \\
\hline
\end{tabular}
\end{table}

The estimated posterior uncertainty continues to provide informative guidance for sample acquisition in this more challenging multifidelity setting. As shown in Fig.~\ref{fig:beam_active_sampling}, regions of high posterior uncertainty overlap with those of large prediction error, enabling uncertainty-guided sample selection. This result indicates that the proposed Bayesian MF-LNO remains capable of identifying informative HF samples even when the LF and HF models exhibit different oscillation frequencies.

Using only eleven HF trajectories, the proposed Bayesian MF-LNO accurately reproduces the HF tip-displacement trajectories despite the substantial discrepancy between the two fidelity levels. As shown in Fig.~\ref{fig:solution_ex3}, the proposed method closely matches the reference responses, whereas the HF-LNO, LF-LNO, and MF-DeepONet exhibit noticeably larger prediction errors. This qualitative observation is quantitatively confirmed in Table~\ref{tab:trajectory_error_comparison_beam}, where the proposed method consistently achieves the lowest relative $L_2$ error, RMSE, and MAE across all representative test trajectories. As additional HF trajectories are acquired, the prediction errors decrease consistently, with uncertainty-guided acquisition achieving faster error reduction than random acquisition, as shown in Fig.~\ref{fig:beam_error_comparison}. The same trend is reflected in Table~\ref{tab:beam_error_comparison_compact}, where the proposed Bayesian MF-LNO consistently attains the lowest average and maximum relative $L_2$ errors for all HF sample sizes.

\section{Conclusion}
\label{sec_discuss}
This work proposed a Bayesian MF-LNO for uncertainty-guided active learning of parametric oscillatory PDEs. The proposed method estimates predictive uncertainty using reSGLD and iteratively acquires informative HF samples for multifidelity operator learning.

The numerical experiments demonstrate that the proposed Bayesian MF-LNO consistently improves data efficiency for multifidelity operator learning through uncertainty-guided HF sample acquisition. In the Lorenz system (Experiment~\ref{Sec:ex1}), the proposed method accurately reproduces the oscillatory dynamics using only nine HF trajectories while consistently achieving lower prediction errors than the baseline MF-LNO and MF-DeepONet. In the Duffing oscillator (Experiment~\ref{Sec:ex2}), the proposed uncertainty-guided acquisition strategy further improves prediction accuracy and enables accurate surrogate construction from only a small number of HF trajectories. In the composite beam problem (Experiment~\ref{Sec:ex3}), the proposed method remains effective despite the more challenging multifidelity setting in which the LF and HF models exhibit different oscillation frequencies, consistently achieving the lowest prediction errors among all competing methods.
A limitation of the proposed Bayesian MF-LNO is that the reliability of the uncertainty indicator depends on the reSGLD hyperparameter configuration as depicted in Fig.~\ref{fig:lorenz_correlation_sensitivity}.

Several directions remain for future research. First, the current framework considers only two fidelity levels, whereas extending the method to hierarchical multifidelity settings with more than two fidelity levels would broaden its applicability. Second, although reSGLD provides informative uncertainty estimates, Bayesian posterior sampling remains computationally demanding, motivating the exploration of alternative Bayesian inference methods with improved scalability. 
Finally, learning LF--HF discrepancies in a shared latent space, rather than directly in the solution space, is a promising direction for future work.

\section*{Acknowledgment}
This work is supported by Sandia National Laboratories (SNL, Contract No. DE-NA0003525) under the DOE Interlaboratory (IL) Laboratory Directed Research and Development (LDRD) Pilot Program, the National Science Foundation under grants DMS-2053746, DMS-2134209, ECCS-2328241, CBET-2347401, and OAC-2311848. The U.S. Department of Energy also supports this work through the Office of Science Advanced Scientific Computing Research program, under the ``Uncertainty Quantification for Multifidelity Operator Learning (MOLUcQ)'' project (Project No. 81739), the SciDAC LEADS Institute (DE-SC0023161), and the Office of Fusion Energy Sciences (DE-SC0024583). Any subjective views or opinions that might be expressed in the paper do not necessarily represent the views of the U.S. Department of Energy or the United States Government.

% \section*{Data Availability}
% The dataset used in this paper will be publicly available after publication. Data can be downloaded from \href{http://github.com/haoyangzheng1996/}{github.com/haoyangzheng1996}.

\section*{Code Availability}
The code used in this study is available from the corresponding author upon reasonable request.
%Code will be released 
% at \href{http://github.com/haoyangzheng1996}{github.com/haoyangzheng1996} 
%after publication.

\bibliography{refs}

\appendix
\section{Ablation Studies}

We include a set of ablation studies to examine how several design choices affect the performance of the proposed Bayesian MF-LNO. Since these tests are supplementary to the main experiments, we focus on the one-dimensional Lorenz and Duffing examples, which already provide representative trends while keeping the discussion concise. In each ablation test, all hyperparameters other than the one under consideration are fixed, and performance is evaluated by the validation or test loss of the resulting MF-LNO model. To reduce the influence of random initialization and stochastic optimization, each configuration is repeated 10 times, and we report the corresponding mean and standard deviation.

A practical issue in these ablation studies is that unsuitable hyperparameter choices may lead to unstable training. In particular, the stochastic noise introduced by reSGLD can amplify this effect and occasionally cause divergence. For this reason, we set the reSGLD temperature to zero in the ablation runs. Since the purpose of this appendix is to identify stable and effective hyperparameter ranges rather than to assess predictive uncertainty, this simplification allows a cleaner comparison across configurations.

\begin{figure}[!htbp]
    \centering
    \includegraphics[width=0.8\linewidth]{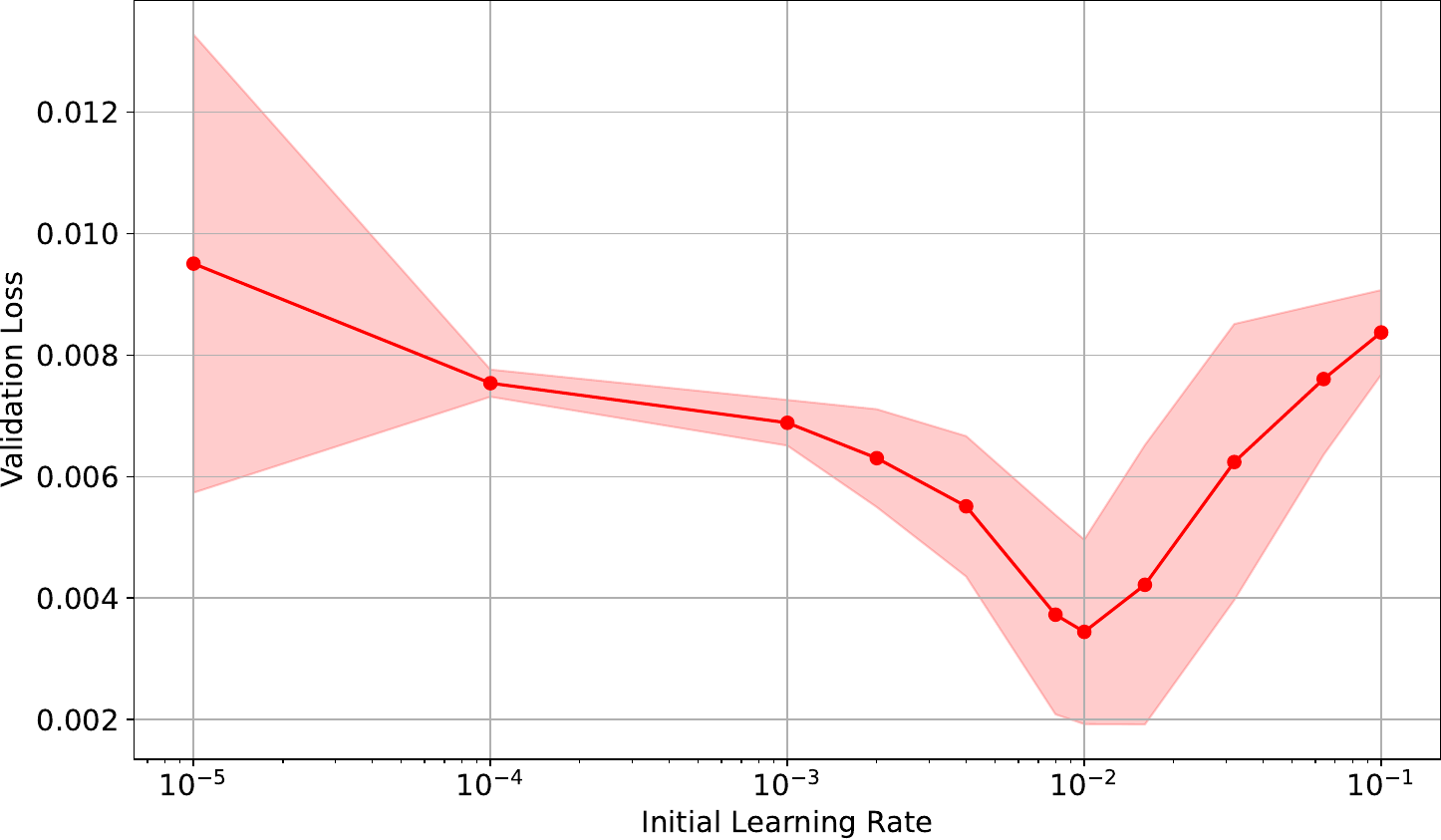}
    \caption{Validation loss versus the initial learning rate for the Duffing oscillator. The shaded region denotes the 95\% confidence interval over repeated runs.}
    \label{fig:duffing_lr}
\end{figure}

\textbf{Initial learning rate:}
For the Duffing oscillator, the training performance is sensitive to the choice of the initial learning rate. To examine this effect, we vary the initial learning rate while keeping the remaining settings fixed, and repeat each configuration 10 times. The LF stage is trained for 2,000 epochs, followed by 5,000 epochs of MF training. A cosine decay schedule is used so that the learning rate decreases to 1\% of its initial value at the end of training.

We first explore a broad range from $10^{-5}$ to $10^{-1}$ and then refine the search within the most promising interval. The results in Fig.~\ref{fig:duffing_lr} indicate that an initial learning rate of $10^{-2}$ gives the lowest validation loss. Based on this observation, we set the smallest- and largest-learning-rate chains in reSGLD as $\eta^{(1)} = 5 \times 10^{-3}$ and $\eta^{(N_C)} = 5 \times 10^{-2}$, respectively.

\textbf{Number of chains in reSGLD:}
We also examine the effect of the number of replica chains in reSGLD using the one-dimensional Lorenz example. Similar trends were observed in the Duffing problem, so we report only the Lorenz results here for brevity. In this study, the smallest and largest learning rates are fixed as $\eta^{(1)}=5 \times 10^{-6}$ and $\eta^{(N_C)}=5 \times 10^{-4}$, and the intermediate learning rates are distributed geometrically between them.

\begin{figure}[!htbp]
    \centering
    \includegraphics[width=0.8\linewidth]{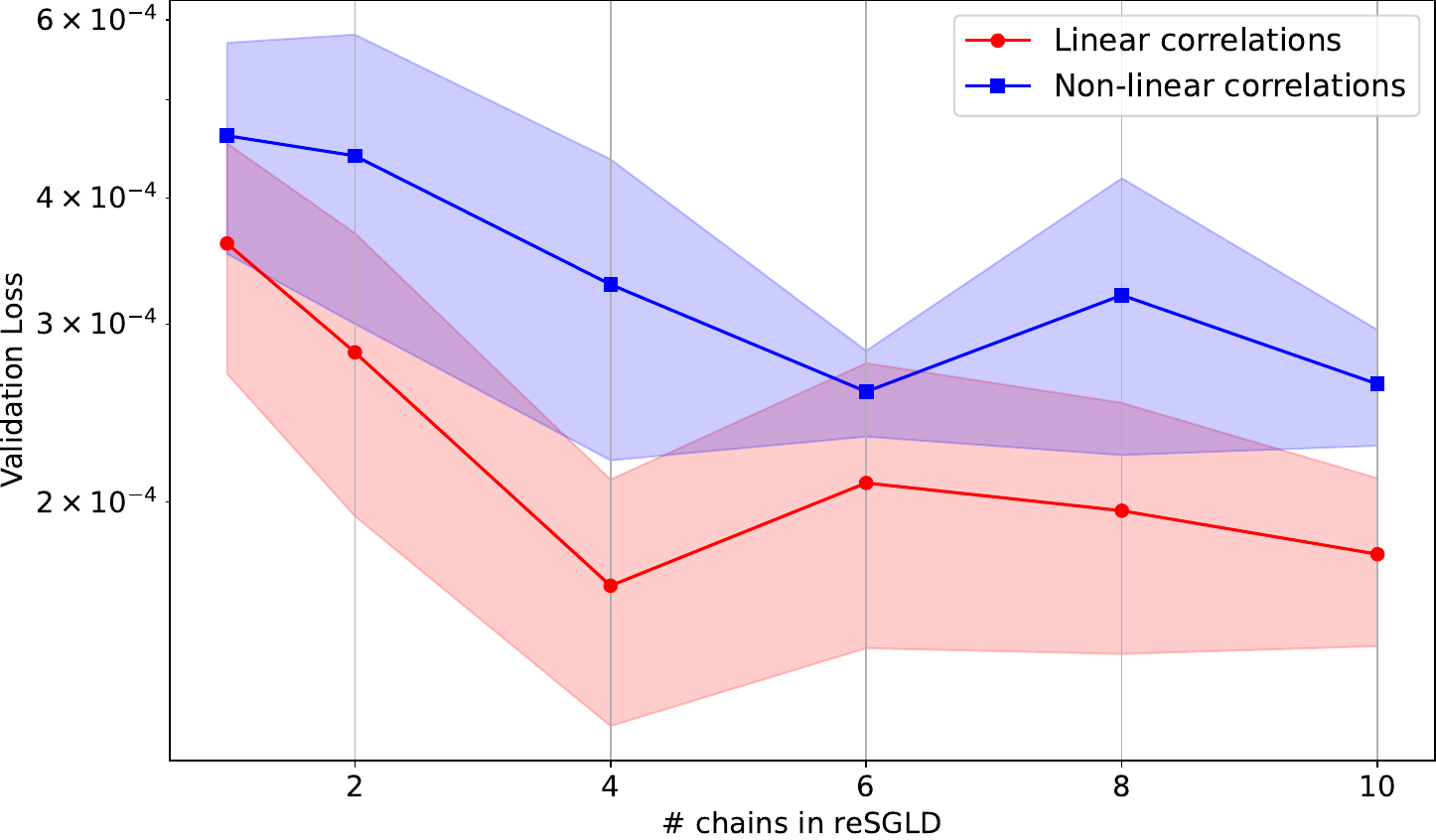}
    \caption{Validation loss versus the number of replica chains in reSGLD for the Lorenz system. The red and blue curves correspond to linear and nonlinear inter-fidelity relations, respectively, and the shaded regions denote the 95\% confidence intervals.}
    \label{fig:ablation_chains}
\end{figure}

The performance is evaluated for $N_C = 1, 2, 4, 6, 8, 10$, considering both linear and nonlinear inter-fidelity relations. Each configuration is repeated 10 times after 2,000 LF-training epochs and 5,000 MF-training epochs. The results are shown in Fig.~\ref{fig:ablation_chains}.

The results show that the optimal number of chains depends on the complexity of the inter-fidelity relation. Linear relations achieve their best performance with fewer chains, whereas nonlinear relations benefit from a moderately larger number of chains. In our experiments, $N_C=6$ provides a good balance between accuracy and computational cost for the nonlinear setting.

\end{document}